\documentclass[a4paper,fleqn]{cas-dc}
\usepackage{microtype}
\usepackage[numbers, square]{natbib}  
\usepackage{graphicx}  
\usepackage{changepage}
\usepackage{subfiles}
\usepackage{IEEEtrantools}
\usepackage{tikz}
\usepackage[edges]{forest}
\usepackage{float}     
\usepackage{xcolor}
\usepackage{xcolor}%
\usepackage{textcomp}%
\usepackage{manyfoot}%
\usepackage{booktabs}%
\usepackage{algorithm}%
\usepackage{algorithmicx}%
\usepackage{algpseudocode}%
\usepackage{listings}%
\usepackage{changepage}
\usepackage{threeparttable}
\usepackage{colortbl}
\usepackage{subfiles}
\usepackage{IEEEtrantools}
\usepackage{tikz}
\usepackage{url}
\usepackage[edges]{forest}
\usepackage{float}     
\usepackage{xcolor}
\usepackage{dashrule}
\usepackage{tcolorbox}
\def\tsc#1{\csdef{#1}{\textsc{\lowercase{#1}}\xspace}}
\tsc{WGM}
\tsc{QE}
\tsc{EP}
\tsc{PMS}
\tsc{BEC}
\tsc{DE}
\begin{document}
\let\WriteBookmarks\relax
\def\floatpagepagefraction{1}
\def\textpagefraction{.001}

\shorttitle{Brain-inspired Computing Based on Deep Learning for Human-computer Interaction: A Review}


\title [mode = title]{Brain-inspired Computing Based on Deep Learning for Human-computer Interaction: A Review}              

\nonumnote{All authors contributed equally to this work.}
\affiliation[inst1]{organization={Shenyang Institute of Computing Technology, Chinese Academy of Sciences},
    city={Shenyang},
    postcode={110168}, 
    country={China}}

\affiliation[inst2]{organization={University of Chinese Academy of Sciences},
    city={Beijing},
    postcode={100049}, 
    country={China}}

\affiliation[inst3]{organization={Heilongjiang Academy of Sciences Intelligent Manufacturing Institute},
    city={Harbin},
    postcode={150090}, 
    country={China}}





\author[inst1,inst2]{Bihui Yu}
\ead{yubihui@sict.ac.cn}

\cortext[corcorr]{Corresponding author.}
\author[inst1,inst2]{Sibo Zhang}
\ead{zhangsibo22@mails.ucas.edu.cn}
\cormark[1]

\author[inst3]{Lili Zhou}
\ead{Zhoulilivip@126.com}

\author[inst1,inst2]{Jingxuan Wei}
\ead{weijingxuan20@mails.ucas.edu.cn}

\author[inst1,inst2]{Linzhuang Sun}
\ead{sunlinzhuang21@mails.ucas.ac.cn}

\author[inst1,inst2]{Liping Bu}
\ead{buliping@sict.ac.cn}


\begin{abstract}
The continuous development of artificial intelligence has a profound impact on biomedicine and other fields, providing new research ideas and technical methods. Brain-inspired computing is an important intersection between multimodal technology and biomedical field. Focusing on the application scenarios of decoding text and speech from brain signals in human-computer interaction, this paper presents a comprehensive review of the brain-inspired computing models based on deep learning (DL), tracking its evolution, application value, challenges and potential research trends. We first reviews its basic concepts and development history, and divides its evolution into two stages: recent machine learning and current deep learning, emphasizing the importance of each stage in the research of brain-inspired computing for human-computer interaction. In addition, the latest progress of deep learning in different tasks of brain-inspired computing for human-computer interaction is reviewed from five perspectives, including datasets and different brain signals, and the application of key technologies in the model is elaborated in detail. Despite significant advances in brain-inspired computational models, challenges remain to fully exploit their capabilities, and we provide insights into possible directions for future academic research. For more detailed information, please visit our GitHub page:\href{https://github.com/ultracoolHub/brain-inspired-computing}{https://github.com/ultracoolHub/brain-inspired-computing}.
\end{abstract}


\maketitle

\section{INTRODUCTION}

Brain-inspired intelligence is a kind of machine intelligence which is inspired by neural mechanism and cognitive behavior mechanism by means of computational modeling and realized by software and hardware cooperation. Brain-inspired intelligence system is brain-inspired in information processing mechanism and human-like in cognitive behavior and intelligence level. Human brain activity is a complex and continuous dynamic process, and its complexity is far beyond the upper limit that can be simulated by current computing resources, so people have not given up the exploration of the brain. Brain-inspired computing is founded upon the structural framework and operational principles of the human brain, and integrates the current computational development path of computer science and neuroscience\cite{furber2016brain,zhang2020system,parhi2020brain,mehonic2022brain}.

Researchers are constantly trying to understand the neural mechanisms and cognitive behavior through the study of the brain. Traditional DL necessitates an extensive collection of annotated datasets for effective training, and manual annotated data is expensive and affected by human subjective consciousness, so the annotation results are not completely accurate. In contrast, the brain weighs about 2.5 pounds and consumes only about 40\% to 60\% of the body's blood sugar\cite{squire1988memory}. If people can use their own physiological data to decode text or speech and complete codec tasks similar to machine translation, it can not only save manpower, but also have important significance and value for the cognitive mechanism and cognitive ability of the human brain.
\definecolor{hidden-draw}{RGB}{0,0,255} 
\definecolor{hidden-pink}{RGB}{1.0, 0.41, 0.71}
\tikzstyle{my-box}=[
    rectangle,
    rounded corners,
    text opacity=1,
    minimum height=1.5em,
    minimum width=5em,
    inner sep=2pt,
    align=center,
    fill opacity=.5,
    line width=0.8pt,
]
\tikzstyle{leaf}=[my-box, minimum height=1.5em,
    fill=hidden-pink!80, text=black, align=left,font=\normalsize,
    inner xsep=2pt,
    inner ysep=4pt,
    line width=0.8pt,
]
\begin{figure*}[!th]
     \begin{adjustwidth}{2em}{-2em}  
        \resizebox{0.94\textwidth}{!}{  
        \begin{forest}
            forked edges,
            for tree={
                grow=east,
                reversed=true,
                anchor=west,
                parent anchor=east,
                child anchor=west,
                base=left,
                font=\large,
                rectangle,
                draw=hidden-draw,
                rounded corners,
                align=left,
                minimum width=4em,
                edge+={darkgray, line width=1pt},
                s sep=5pt,
                inner xsep=4pt,
                inner ysep=1pt,
                line width=0.8pt,
                ver/.style={rotate=90, child anchor=north, parent anchor=south, anchor=center},
            },
            where level=1{text width=5em,fill=orange!10}{},
            where level=2{text width=5em,fill=blue!10}{},
            where level=3{yshift=0.26pt,fill=pink!30}{},
            where level=4{yshift=0.26pt,fill=yellow!20}{},
            where level=5{yshift=0.26pt}{},
            [
               Paper \\
               Structure, fill=green!20,text width=6em
                [
                    Evolution of \\
                    Brain-inspired Computing\\
                    for HCI, text width=13.4em,l=4cm
                    [
                      Recent Models \\
                      Based on\\
                      Machine Learning, text width=10em,l=5em
                    [
                      Classification, text width=10em
                       [
                            NB\cite{webb2010naive}{,}
                            SVM\cite{noble2006support}{,}
                            KNN\cite{peterson2009k}{,}\\
                            LR\cite{cox1958regression}{,}
                            RF\cite{breiman2001random}{,}
                            LDA\cite{balakrishnama1998linear}{,}
                            ,fill=green!0
                        ]
                    ]
                    [
                      Feature Extraction, text width=10em
                        [
                            PCA\cite{wold1987principal}{,}
                            AR\cite{shibata1976selection}{,}
                            FFT\cite{nussbaumer1982fast}{,}
                            WT\cite{farge1992wavelet}
                            ,fill=green!0
                        ]
                     ]
                   ]  
                   [
                     Current Models\\ 
                     Based on \\
                     Deep Learning, text width=10em,l=5em
                     [
                      Early DL Model, text width=10em
                        [
                            CNN\cite{schirrmeister2017deep}{,}RNN\cite{alhagry2017emotion,sakhavi2018learning}
                            ,fill=green!0
                        ]
                     ]
                     [
                     Pretrained Model, text width=10em
                        [
                            BERT\cite{zou2022cross}{,}BART\cite{wang2022open,feng2023aligning,duan2023dewave}
                            ,fill=green!0
                        ]
                     ]
                  ]
                ]
                [                
                    Application of \\ Deep Learning-based \\ Brain-inspired Computing \\Models for HCI Tasks, text width=13.4em,l=4cm
                    [                     
                        Data aquisition \\ 
                        and preprocessing, text width=10em
                        [
                            Data Acquisition, text width=10em
                            [
                            1. Experimental Design.\\
                            2. EEG Acquisition.\\
                            3. FMRI Acquisition.\\
                            4. Eye-tracking Acquisition.\\
                            ,fill=green!0
                            ]                       
                        ]
                        [
                            Data Preprocessing, text width=10em
                            [
                            1. Filter processing.\\
                            2. Artifact removal.\\
                            3. Reference reset.\\
                            4. Spatial correction.\\
                            5. Time-frequency analysis.\\
                            ,fill=green!0
                            ]                       
                        ]
                        [
                            Public Dataset, text width=10em
                            [
                            ZuCo\cite{hollenstein2018zuco}{,}
                            ZuCo2.0\cite{hollenstein2019zuco}{,}\\
                            Science\cite{mitchell2008predicting}{,}
                            NC\cite{pereira2018toward}
                            ,fill=green!0
                            ]
                        ]
                    ]
                    [
                        Models based \\
                        on EEG , text width=10em
                        [
                            Recent Progress, text width=10em
                            [
                            Automatic Speech Recognition\cite{krishna2019state}{,}\\
                            EEGNet\cite{cooney2020evaluation}{,}\\
                            EEG-transformer\cite{lee2022eeg}{,}\\
                            EEG-to-text\cite{wang2022open}{,}\\
                            CSCL\cite{feng2023aligning}{,}
                            DeWave\cite{duan2023dewave}
                            ,fill=green!0
                            ]
                        ]
                        [
                            Key Technologies, text width=10em
                            [
                             1. EEG-to-text Decoding.\\
                             2. Curriculum Learning.\\
                             3. Discrete Codex.
                             ,fill=green!0
                            ]
                        ]
                    ]
                    [
                        Models based \\
                        on fMRI, text width=10em
                        [
                            Recent Progress, text width=10em
                            [
                            Context Into Language\cite{jain2018incorporating}{,}\\
                            Brain2word\cite{affolter2020brain2word}{,}\\
                            multi-timescale models\cite{jain2020interpretable}{,}\\
                            Semantic-based Classification\cite{lin2022neural}{,}\\
                            Self-supervised Learning\cite{millet2022toward}{,}\\
                            Cross-madal Cloze Task\cite{zou2022cross}{,}\\
                            UniCoRN\cite{xi2023unicorn}{,}\\
                            Semantic Reconstruction\cite{tang2023semantic}
                            ,fill=green!0
                            ]
                        ]
                        [
                            Key Technologies, text width=10em
                            [
                            1. Brain Decoding Model.\\
                            2. Semantic Features Fusion.
                            ,fill=green!0
                            ]
                        ]
                    ]
                     [
                        Models based \\
                        on MEG, text width=10em
                        [
                            Recent Progress, text width=10em
                            [
                            Seq2seq Learning\cite{dash2019automatic}{,}\\
                            Decoding Speech\cite{dash2019decoding}{,}\\
                            Decoding Imagined and Spoken Phrases\cite{dash2020decoding}{,}\\
                            MEG Sensor Selection\cite{dash2020meg}{,}\\
                            Decoding Speech\cite{defossez2023decoding}
                            ,fill=green!0
                            ]
                        ]
                        [
                            Key Technologies, text width=10em
                            [
                            1. CNNs And Transfer Learning.\\
                            2. Signal Alignment.
                            ,fill=green!0
                            ]
                        ]
                    ]
                    [
                        Models based \\
                        on ECoG, text width=10em
                        [
                            Recent Progress, text width=10em
                            [
                            Machine Translation\cite{makin2020machine}{,}\\
                            Brain2char\cite{sun2020brain2char}{,}\\
                            Neural Speech Decoding\cite{chen2023neural}{,}\\
                            Direct Speech Reconstruction\cite{berezutskaya2023direct}{,}\\
                            Synthesizing Speech\cite{shigemi2023synthesizing}
                            ,fill=green!0
                            ]
                        ]
                        [
                            Key Technologies, text width=10em
                            [
                            1. Decoding Approach.\\
                            2. Neural Vocoder.
                            ,fill=green!0
                            ]
                        ]
                    ]
                ]
                [
                  Challenges and \\ Future Directions 
                  of \\ Deep Learning-based\\
                  Brain-inspired Computing \\ Models for HCI Tasks, text width=13.4em,l=4cm
                  [
                    Challenges, text width=10em
                    [
                    Model Training, text width=10em
                    [
                    1. Challenges for Brain Dataset.\\
                    2. Computational Resource Demand.\\
                    3. Acuracy and Generalization of the Model.
                    ,fill=green!0
                    ]
                    ]
                    [
                    Model Application, text width=10em
                    [
                    4. Challenges for Real-time Processing.\\
                    5. Challenges for BCI Technology.
                    ,fill=green!0
                    ]
                    ]
                    [
                    Ethical Challenges, text width=10em
                    [
                    6. Challenges for Ethical Concerns.
                    ,fill=green!0
                    ]
                    ]
                  ]
                  [
                    Future Directions, text width=10em
                    [
                        1. Landing Application of Brain-inspired Computing.\\
                        2. BCIs across Language Boundaries.\\
                        3. Spiking Nerual Network(SNN).\\
                        ,fill=green!0
                    ]
                  ]
                ]
            ]
        \end{forest}
}
    \end{adjustwidth}
\caption{An introduction to the deep learning-based brain-inspired computing models for HCI tasks, including its history, currently available methods and techniques, as well as the challenges encountered and potential solutions, and the direction of future development.}
\centering
\label{fig:1}
\end{figure*}
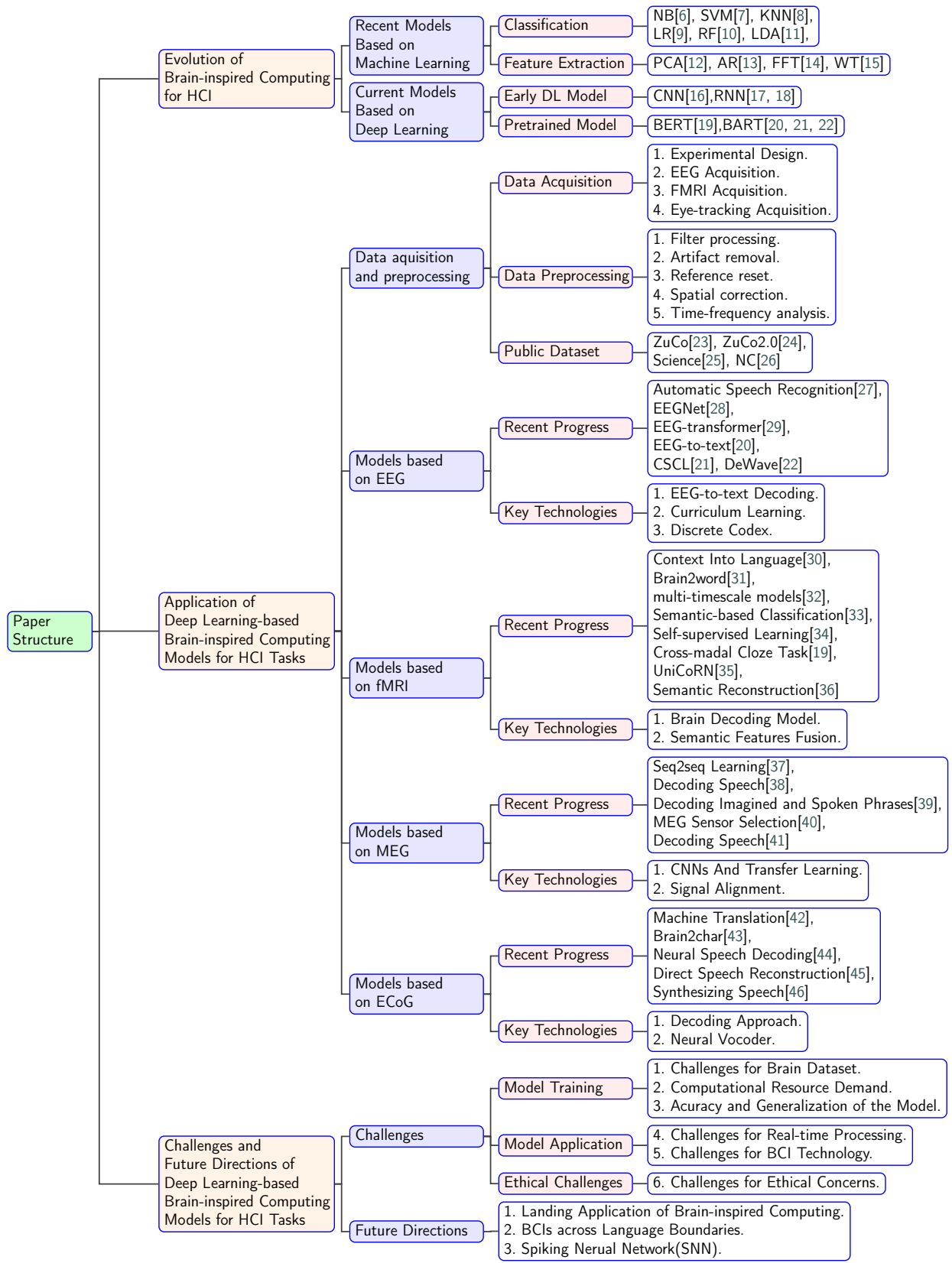

The continuous exploration of the brain forms the basis of brain-inspired computing, which not only draws inspiration from the complex structure and operation principle of the brain, but also focuses on the innovative use of physiological data of brain signals for practical applications. Brain-inspired computing models can be applied in human-computer interaction(HCI). And a typical application is brain-computer interface(BCI) which aims to establish a channel between the brain and the external environment, which does not depend on the peripheral nervous system, and realizes the information exchange and control between the brain and external devices with processing or computing capabilities. Through the acquisition, analysis and processing of  brain signals, human can directly interact with computers without relying on external devices. For example, the main goal of cognitive BCI is to understand and analyze the processing mechanism of speech information in human brain. It can help patients overcome the difficulties in language expression and text input by recognizing the ideas expressed by brain signals and converting them into speech or text output\cite{daly2008brain}.

As an emerging research field, the exact definition of brain-inspired computing is still unclear. This field covers computational theory, architecture design, hardware specifications, etc., while learning from brain information processing mechanisms and biological physiological structures to build a variety of models and algorithms\cite{qu2022review}. Some comprehensive studies systematically review the research progress in the field of brain-inspired computing from multiple perspectives. Some studies focus on brain-inspired computing in a narrow sense, such as spike-based neural mimic computing in vision applications\cite{hendy2022review}. At the same time, several studies have comprehensively reviewed hybrid neural networks, highlighting their comprehensive balance with respect to artificial neural networks and spiking neural networks\cite{liu2024advancing}. There are also studies that explore cognitive engineering approaches, aiming to assist the research community in designing more consistent methods and techniques to advance cognitive machines\cite{son2016survey}. In addition, some reviews focus on the hardware design of brain-inspired computing architectures\cite{park2023complex}. On the other hand, some reviews start from the challenges faced by the field of brain-inspired computing and propose a general framework for brain-inspired computing systems in practical applications\cite{li2023brain}. There are also some research from the perspective of the application of BCI systems, but only summarize the signal processing and calculation methods related to EEG signal\cite{gu2021eeg,kalagi2017brain}.

These reviews provide valuable insights into the progress of brain-inspired computational models. Inspired by these reviews, this paper deeply investigates the development of brain-inspired computing models from the perspective of machine learning(ML) and deep learning(DL), focusing on the brain-inspired computing models that decode text or speech through brain signals in human-computer interaction. This paper summarizes the application scenarios and limitations of different machine learning algorithms in brain-inspired computing models, and comprehensively reviews the deep learning related models of brain-inspired computing involving EEG, fMRI, MEG, ECoG in recent years. At the same time, the application of key technologies in these models is described in detail. This paper aims to summarize the enlightenment of artificial neural networks in brain-inspired computing models and algorithms in human-computer interaction scenarios from the perspective of generalized brain-inspired computing, expecting to fill the gap in the review of brain-inspired computing models for human-computer interaction based on deep learning. The contributions of this review can be summarized in three main points:
\begin{itemize}
    \item To enhance understanding of deep learning-based brain-inspired computing models for human-computer interaction, we first introduce the fundamental concepts. Then, we review recent advancements in brain-inspired computing within machine learning and deep learning, highlighting the critical role of machine learning in feature extraction and classification, and leading to the primary focus on deep learning-based brain-inspired computing models.
    \item After reviewing the progress of the research, we divide the tasks related to brain-inspired computing for human-computer interaction based on deep learning models into five parts, four of which introduce the progress based on different brain signals, emphasize the importance of experimental stimulus design and data acquisition and preprocessing, and elaborate the application of key technologies in the model.
    \item In addition, we highlight the challenges faced by brain-inspired computational models based on deep learning from three aspects: model training, model application, and ethical issues, and propose potential solutions. We look forward to future applications such as cross-lingual BCI and spiking neural network with greater biological plausibility.
\end{itemize}

This paper will organize the rest following the structure shown in Figure \ref{fig:1}: we will first delve into the basic concepts and development history related to brain-inspired computing, brain-computer interfaces, brain signals and so on. In section \ref{section2} we will promote a deeper understanding of this research area. Then, in section \ref{section3}, this paper reviews the research results, progress and some key technologies of researchers using deep learning models and algorithms to study brain signal data through a series of specific case studies to illustrate the practical effects. In section \ref{section4}, we will deeply explore the challenges and limitations of DL models in brain-inspired computing for human-computer interaction tasks, such as data accuracy requirements and computational resource requirements, as well as the limitations and security risks of BCI systems. Future research directions such as SNN and cross-lingual BCI systems are discussed in section \ref{section5}. This paper concludes the entire discussion in section \ref{section6}.

\section{BASIC CONCEPTS AND EVOLUTION}
\label{section2}

This section mainly expounds the related concepts of brain-inspired computing, brain-computer interface and pre-training model, and focuses on the common algorithms of machine learning in brain-inspired computing for human-computer interaction tasks, as well as the important applications of deep learning models.

\subsection{Basic Concepts of Brain-inspired Computing for HCI}
Brain-inspired computing commonly encompasses computational theories, architectural designs, hardware specifications, and various models and algorithms that draw inspiration from the information processing mechanisms and biophysiological structures of the brain\cite{qu2022review}. This review studies brain-inspired computing from a broad perspective. Learning from the structure and working principle of the brain, but not limited to the simulation of the brain, but also including the integration of traditional artificial neural network with more brain-inspired characteristics of heterogeneous neural networks, is an integration of current computer science and neuroscience Computing development path. This paper focuses on the brain-inspired computing models and algorithms for decoding text and speech from brain signals in HCI tasks through ML and DL.

In order to better understand the task of brain signal decoding, we will briefly discuss the various brain signals that are commonly used, including electroencephalogram(EEG), functional magnetic resonance imaging(fMRI), Magnetoencephalography(MEG) and electrocorticography(ECoG).

EEG, a non-invasive method for measuring brain activity, encapsulates a rich repository of information regarding the intricate workings of the brain, rendering it an appealing data source for applications in DL. The advancement of DL techniques in the analysis of EEG features holds tremendous potential for the development of novel diagnostic and therapeutic tools within the realm of neurological disorders. Typically, EEG signals are acquired through the placement of electrodes on the scalp and are represented as a two-dimensional matrix or graph. This representation unfolds in time as one dimension and spatially in terms of electrode locations as another. Researchers in the field commonly treat multiple channels as the spatial dimension of EEG data, aligning with the temporal dimension\cite{hollenstein2019cognival,wang2023large}.

FMRI is a another powerful non-invasive tool for studying brain function\cite{smith2004overview}. Used for psychologists, psychiatrists, and neurologists, fMRI provides high-quality visualizations of brain activity in response to sensory stimulation or cognitive functions. Thus, it facilitates the study of the intricate workings of a healthy brain\cite{wang2022open,smith2004overview}.The objective of fMRI analysis is to robustly, sensitively, and validly detect the regions of the brain that exhibit heightened intensity during the moments when stimulation is applied\cite{feng2023aligning}.

MEG is a non-invasive neuroimaging technique used to record neural activity in the brain. It reflects brain activity by measuring the weak magnetic fields generated by neurons in the brain. These magnetic fields are able to penetrate the skull and tissue without being affected by them, thus providing EEG signals with high spatio-temporal resolution. In MEG, an array of induction coils is placed around the patient's head to detect and record changes in the magnetic field caused by neuronal activity. Because MEG measures the magnetic fields generated within the brain by nerve currents, rather than the electrical potentials (such as EEG) on the scalp surface, it provides a more precise spatial resolution. MEG is often used to study functional localization of the brain, neuroplasticity, and brain activity associated with neurological diseases. Its non-invasive nature makes it an important tool for studying brain function and abnormal activity. MEG is increasingly used in neuroscience, clinical medicine and cognitive research, providing a powerful means to deeply understand the complex functions of the brain.

The three non-invasive methods described earlier, as a contrast, ECoG is a technique that records electrical activity in the brain, but electrodes are implanted directly on the surface of the patient's brain rather than placed on the scalp. By implanting an array of electrodes on the surface of the brain\cite{makin2020machine}, ECoG is able to provide EEG signals with higher spatio-temporal resolution. This method is commonly used to study epilepsy, neuroplasticity, and brain function localization. Compared to traditional electroencephalograms, ECoG provides a finer spatial resolution, allowing researchers to understand the electrical activity of specific brain regions in greater detail.

The above summarizes the commonly used brain signals. The improvement of deep learning algorithms, especially the development of pre-trained models, has promoted the performance improvement of brain signal decoding tasks in the field of brain-inspired computing. Most methods of processing brain physiological data with deep learning techniques are combined with the currently hot pre-trained language model (PLM) which has undergone a series of evolution. As an early initiative, ELMo\cite{peters-etal-2018-deep} aimed to capture context-aware word representations through the pre-training of a bidirectional long short-term memory network (biLSTM). Unlike learning fixed word representations, ELMo's approach involves fine-tuning the biLSTM network for specific downstream tasks. Subsequently, drawing on the highly parallelized transformer architecture\cite{vaswani2017attention} and the self-attention mechanism, BERT\cite{devlin2018bert} was introduced. BERT entails pre-training two-way language models and specifically designed pre-training tasks using a large-scale unlabeled corpus. These pre-trained, context-aware word representations serve as highly effective, generic semantic features, significantly elevating the performance standard for a wide range of natural language processing tasks. This study has spurred a substantial amount of subsequent research, establishing a prevalent "pre-train and fine-tune" learning paradigm. Following this paradigm, a great deal of research has been done on PLMs, introducing different architectures such as generative models GPT-2\cite{radford2019language} and BART\cite{lewis2019bart}, in which PLM often needs to be fine-tuned to suit different downstream tasks.

\subsection{The Evolution of Brain-inspired Computing for HCI}

With the rapid development of artificial intelligence models, brain-inspired computing models for HCI have also attracted wide attention. Brain-inspired computing tasks that can be completed are increasingly complex from early machine learning models to large-scale deep learning models with hundreds of billions of parameters.

\subsubsection{Recent Models Based on Machine Learning}

\begin{table*}[htbp]
\footnotesize
  \centering
  \caption{Machine learning algorithms used in brain-inspired computing tasks.}
  \resizebox{\linewidth}{!}{ 
    \begin{tabular}{ccr}
    \toprule
    \multicolumn{1}{c}{Task} & \multicolumn{1}{c}{Stage Category} & \multicolumn{1}{l}{Representative Work} \\
    \midrule
        \multirow{10}[2]{*}{Model Based on Machine Learning} & \multirow{6}[1]{*}{Classification} &  \multicolumn{1}{l}{NB, \cite{webb2010naive}} \\
          &       & \multicolumn{1}{l}{SVM, \cite{noble2006support}} \\
          &       & \multicolumn{1}{l}{KNN, \cite{peterson2009k}} \\
          &       & \multicolumn{1}{l}{LR, \cite{cox1958regression}} \\
          &       & \multicolumn{1}{l}{RF, \cite{breiman2001random}} \\
        &       & \multicolumn{1}{l}{LDA, \cite{balakrishnama1998linear}} \\
        \cmidrule{2-3} 
          & \multirow{4}[0]{*}{Feature Extraction} & \multicolumn{1}{l}{PCA,
          \cite{wold1987principal}} \\
          &       & \multicolumn{1}{l}{AR, \cite{shibata1976selection}} \\   
          &       & \multicolumn{1}{l}{FFT, \cite{nussbaumer1982fast}} \\
          &       & \multicolumn{1}{l}{WT, \cite{farge1992wavelet}} \\
    \bottomrule
    \end{tabular}%
    }
  \label{table1}%
\end{table*}%

In recent years, the use of machine learning techniques to analyze brain signals has attracted a lot of attention. For example, there is growing research evidence that machine learning can extract meaningful information from high-dimensional and noisy EEG signals. Given the interest and widespread use of the technology, this section focuses on recent examples of researchers using machine learning to analyze brain signals to build brain-inspired computational models, including machine learning methods and related applications. As shown in Table \ref{table1}, two main applications of machine learning methods in the field of brain-inspired computing at different stages, namely classification and feature extraction, are shown, and common machine learning algorithms are listed.

\begin{figure*}[htbp]
    \centering
    \includegraphics[width=0.7\textwidth]{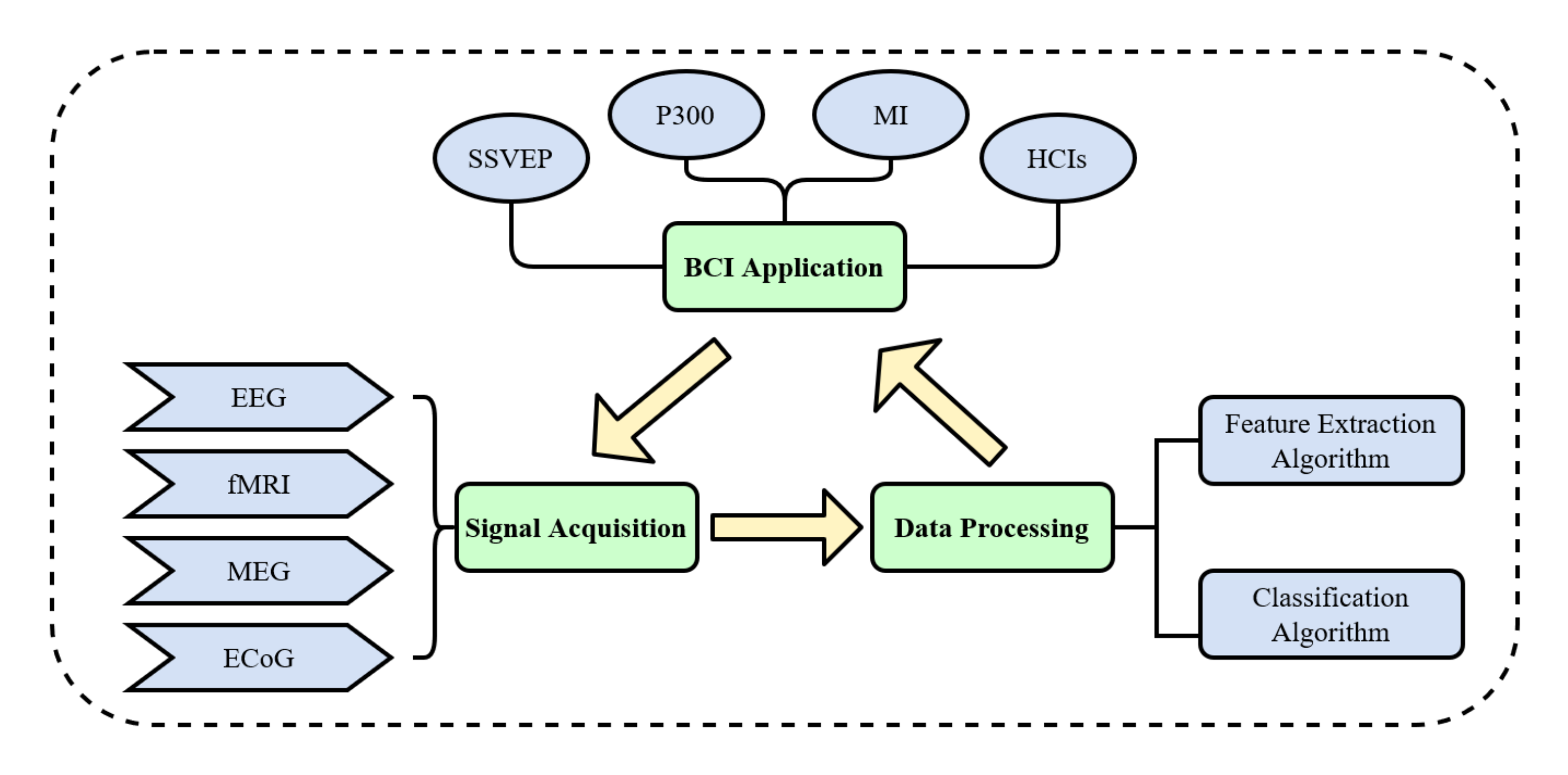}
    \caption{Diagram of the BCI system module. }
    \label{fig:0}
\end{figure*}

Among the applications of brain-inspired computing for human-computer interaction, brain-computer interface is the most widespread one. Brain-computer interface technology establishes a connection between the human brain and the computer and its external equipment for information exchange. As shown in Figure \ref{fig:0}, a complete brain-computer interface system generally includes three modules of brain signal acquisition, data processing and BCI application, in which different signals such as EEG, fMRI, MEG and ECoG can be selected according to the actual situation. Data processing will combine feature extraction algorithms and classification algorithms to further enhance the collected brain signals, and then realize information exchange with external devices through brain computer interface applications\cite{gurkok2012brain,mridha2021brain}. The typical applications include SSVEP\cite{zhang2015ssvep,luo2010user}, P300\cite{fazel2012p300,fouad2020improving}, MI\cite{hamedi2016electroencephalographic,jin2019correlation} and some other human-computer interaction application scenarios.

Among the above mentioned data processing algorithms, machine learning algorithms are used to solve the problem in the earliest brain-inspired computing models.

On the one hand, machine learning algorithms are commonly employed in various traditional classification approaches, encompassing both supervised and unsupervised learning. Among these, supervised algorithms stand out as the most renowned method in EEG data analysis\cite{fiscon2018combining}. It mainly includes naive bayes (NB), support vector machine (SVM), k-nearest neighbor (KNN), logistic regression (LR), random forest (RF) and linear discriminant analysis(LDA). Each supervised model employs a learning algorithm to produce a more accurate model\cite{saeidi2021neural,tan2006classification}. These methods are often used to complete classification tasks in brain-inspired computing.

For example, NB is a probabilistic classifier that can be applied to EEG signal analysis. It only needs a small amount of training data set to classify data according to certain features by Bayes' theorem\cite{rakshit2016naive,hosseini2020review}, so it plays a certain advantage in BCI system\cite{machado2013study,he2015common}. The tasks of applying NB include emotion recognition\cite{oktavia2019human}, epilepsy detection\cite{lestari2020epileptic} and motor imagery\cite{reshmi2013design,wang2016detection,sagee2017eeg,isa2019motor,aggarwal2019signal}. However, NB has certain limitations, under its basic assumption, all properties are independent of each other, and the eigenvector has the same effect on the result\cite{hosseini2020review}.

Next, SVM is widely used in EEG classification and BCI systems\cite{lal2004support,arbabi2006comparison,li2008self,hortal2013online,hou2017improving}. SVM can effectively segregate two datasets, whether in a linear or non-linear fashion. For linear separation, SVM leverages discriminant hyperplanes to delineate classes, while for non-linear separation, it employs kernel functions to discern decision boundaries.  Compared with KNN and other supervised algorithms, SVM has lower computational complexity\cite{palaniappan2014comparative,osowski2004mlp}. SVM have found extensive application in EEG signal classification owing to their simplicity and versatility in addressing classification challenges, including the diagnosis of brain diseases\cite{moctezuma2020classification,trambaiolli2011improving,murugavel2016hierarchical}. In addition, SVM can also be used for other tasks, such as robot control\cite{hortal2015svm} and emotion classification\cite{mehmood2015emotion}. However, SVM performance is affected by kernel function and penalty coefficient parameters. Therefore, it is important to optimize the parameters introduced into the SVM classifier\cite{du2015adaptive}. 

Furthermore, the basic idea of KNN algorithm is that in feature space, if most of the neighbors of an instance belong to a certain class, then the instance also belongs to that class\cite{peterson2009k}. KNN algorithm has strong interpretability, easy implementation, and no parameter adjustment, only need to select a suitable K value, so it is also used in the classification task of BCI system\cite{yazdani2009classification,rajini2011classification,awan2016effective,yudhana2020human}. However, KNN algorithm also has some limitations. The algorithm is required to compute the distance between each sample and all training samples, so it requires a large amount of computation on large-scale data sets, and KNN algorithm is sensitive to noisy data, which may be greatly affected, resulting in inaccurate prediction results. Furthermore, despite the decrease in computational complexity with an increase in the value of k for KNN, its classification performance also diminishes\cite{palaniappan2014comparative,beyer1999nearest}.

Next, LR algorithm is a common classification algorithm\cite{cox1958regression}, its core idea is to use logical functions to build a linear model to model the relationship between input features and output labels. Logical functions such as sigmoid can map any real number to the interval [0,1] and can therefore be used to represent probabilities. In order to determine the parameters of the model, it is necessary to define a loss function to measure the degree of fit of the model. LR algorithm is suitable for binary classification problems and has strong interpretability. Thus, LR algorithms are commonly used in classification tasks such as EEG classification\cite{ryali2010sparse,zeng2015optimizing,tomioka2006logistic} and BCI systems\cite{fiebig2016multi,siuly2016cross,miladinovic2020slow}. However, it also has some limitations, is susceptible to outliers and redundant features, and is less effective when dealing with multiple classification problems.

Moreover, RF is a model composed of multiple decision trees, each trained with a distinct subset of data and features. It randomly selects data and features, and then integrates the prediction results of each tree to make the final prediction, effectively reducing overfitting and improving the prediction accuracy\cite{ramzan2019learning}. In solving classification problems, RF's parallel structure has better performance than other supervised algorithms in processing large EEG datasets\cite{wang2016detection,edla2018classification,antoniou2021eeg,kumar2021classification,pena2022eeg}. In addition, it is used for image classification\cite{nayak2016brain}, brain tumor detection\cite{anitha2018development} and some other BCI tasks\cite{okumucs2017random}. However, over-fitting and instability of trees can affect the performance of RF models, especially for trees of different sizes\cite{hosseini2020review}.

Lastly, LDA is a supervised learning method designed to identify features that effectively differentiate between different classes. Its fundamental concept involves maximizing inter-class distance while minimizing intra-class distance. By projecting data into a low-dimensional space, it finds the hyperplane that can best distinguish between different categories, thus achieving dimensionality reduction while maximizing classification accuracy\cite{bandos2009classification,tharwat2017linear,lotte2007review}.
LDA has proven successful in addressing classification challenges within BCI systems\cite{gareis2011determination,xu2011enhanced,ishfaque2013evaluation,rashid2019classification}, such as motor imagery\cite{aggarwal2019signal,molla2021trial}, P300 spellers\cite{bostanov2004bci}, brain states decoding\cite{xia2015empirical}, and multi-class BCI\cite{scherer2004asynchronously}, attributed to its straightforward usability and minimal computational demands. 
Nevertheless, the primary limitation of this model lies in its linear nature, rendering it unsuitable for application to non-linear EEG data\cite{beyer1999nearest,garcia2003support}.

On the other hand, in addition to the above classification task algorithms, machine learning also plays an important role in the stage of feature extraction, several feature extraction algorithms based on machine learning are reviewed next, including principal component analysis(PCA), autoregressive(AR), fast fourier transform(FFT) and wavelet transform(WT).

Firstly, the fundamental concept of the PCA algorithm is to transform high-dimensional data into a lower-dimensional space, preserving the main features of the data, while removing redundancy and correlation\cite{wold1987principal}. The PCA algorithm achieves dimensionality reduction through linear transformation, making the data easier to visualize and analyze, while improving the performance and efficiency of machine learning models. Therefore, the algorithm is suitable for a variety of scenarios and can simplify complex problems. In brain-inspired models, PCA algorithm is used for feature extraction\cite{kottaimalai2013eeg,yu2014analysis,vijay2015brain,ilyas2015survey}. However, PCA also has some limitations, for example, it is sensitive to noise and outliers, which may lead to the deviation of dimensionality reduction results from the real situation. In addition, PCA ignores the correlation between samples and does not consider the importance of features, which may affect its comprehensive description of the data. In addition, its interpretation is poor.

Secondly, AR is a prediction algorithm based on time series, the basic idea is to use the observed value of the past time step to predict the current observation value, that is, by establishing a linear model, the data of the current moment and the data of the past moment are linearly combined, so as to obtain the data of the future moment. At the heart of the AR model is the autoregressive coefficient, which represents the relationship between the data at the current moment and the data at the past moment. The autoregressive coefficient can be estimated by least square method to obtain an optimal autoregressive model\cite{shibata1976selection}. AR models capture trends and historical dependencies in time series data. The application of AR algorithm includes feature extraction tasks\cite{wang2010feature}, motor imagery tasks\cite{hettiarachchi2015multivariate} and other BCI tasks\cite{bufalari2006autoregressive,garg2011full,ting2014estimating}.
However, AR model also has some limitations, it can only capture the autoregressive relationship, can not capture the moving average relationship. The AR model ignores the error term of the past time step and may not capture the moving average in the data. For some time series data, the AR model may require a higher order to fit the data well, resulting in an increase in model complexity.

Moreover, FFT algorithm reduces the computational complexity of discrete fourier transform(DFT)\cite{winograd1978computing} algorithm and makes the processing of large-scale data efficient and feasible. Especially for large data sets. This makes it an important tool in signal processing, communication systems and other fields\cite{nussbaumer1982fast}. Applications in BCI systems such as game controlling\cite{djamal2017brain}, feature extraction\cite{amin2022eeg,djamal2017eeg} are commonly used. However, the FFT algorithm demands a substantial amount of memory space for both data storage and result computation. The accuracy of FFT algorithm may decrease for data with non-uniform distribution or noisy data.

Lastly, WT builds upon and enhances the concept of short-time Fourier transform localization, addressing drawbacks like the fixed window size that doesn't adjust with frequency changes. It systematically conducts multi-scale refinement of the signal through successive telescopic translation operations. Consequently, the time subdivision at high frequency and frequency subdivision at low frequency dynamically conform to the demands of time-frequency signal analysis, allowing for focused attention on any signal details. WT can analyze information of different frequency components on the same time scale which can adapt to focus on arbitrary details of the signal which is particularly useful for analyzing non-stationary signals and abrupt signals\cite{farge1992wavelet}. Thus, WT is commonly used in feature extraction\cite{azim2010feature,akin2002comparison,kousarrizi2009feature,al2019feature} and other BCI tasks such as SSVEP\cite{peng2021fatigue,huang2008application,aydemir2011wavelet,mohamed2014enhancing,khalaf2018brain}. However, the computational complexity of the wavelet transform is higher than that of the Fourier transform, and the computation time may be longer for large data sets. Because the boundary of the signal cannot fully satisfy the periodicity condition, there will be an end effect, which may affect the accuracy of the transformation result, and it is necessary to select a suitable wavelet base.

In general, the brain-inspired computing models and algorithms established by machine learning generally have certain limitations. Firstly, the data requirements are high, and the data quality and data quantity have a great impact on the performance of the model. In addition, due to the characteristics of some methods themselves, such as LDA as a representative of linear classification algorithms can not solve the nonlinear data classification problem. LR can well solve the binary classification problem, but can not solve the needs of brain-inspired computing model multi-classification problem. KNN, FFT and other algorithms are sensitive to data noise, the brain signal acquisition itself is easy to be disturbed by the surrounding environment and produce noise. When these machine learning methods are only used to solve brain-inspired computing problems, the effect is often not ideal and the application scenarios are limited. The emergence and rapid development of deep learning algorithms has also promoted the progress of brain-inspired computing models. The advantages of machine learning algorithms in data preprocessing and the advantages of deep learning algorithms in modeling and decoding can be comprehensively used to better deal with brain-inspired computing problems in various HCI application scenarios.


\subsubsection{Current Models Based on Deep Learning}
In recent years, deep learning technology has attracted a lot of attention in several research fields, and can be used to provide a novel solution for learning stable representations in the physiological signals of the brain. The machine learning techniques reviewed earlier can extract valid task-relevant information from signals. Signal classification plays an important role in research tasks and has been applied to many signal control tasks. Although great progress has been made in this regard, there remains considerable potential for enhancing classification accuracy, and there is also a lot of room for exploration in the application of other tasks. 
Hence, as a burgeoning frontier in the realm of machine learning, deep learning has garnered the interest of numerous researchers in the field of brain science.

Based on the Transformer\cite{vaswani2017attention} architecture, BERT\cite{devlin2018bert} learns contextual information forward and backward, enabling the model to better understand the meaning of words in sentences. Pre-trained on large-scale text, BERT learns a common language representation that can then be fine-tuned for various NLP tasks, such as text classification and named entity recognition. BERT model has been applied to brain-inspired computing tasks. For example, the cross-modal cloze task treats the decoding task as a fusion of a direct classification task and a cloze task\cite{zou2022cross}. In natural language processing, cloze tasks have been effectively addressed by BERT, a pre-trained masking language model capable of randomly masking certain words in the input and subsequently predicting the masked words based on context\cite{zou2022cross}. By classifying brain images with context into words from a large number of words, a given context should provide additional information to predict words by narrowing down possible candidates.

BART\cite{lewis2019bart} performs particularly well in the current deep learning pre-training models that combine brain-inspired computing fields. It is unique in its bidirectional and autoregressive learning style. Through a sequence-to-sequence training approach, BART introduces a de-noising mechanism that allows the model to capture global text information by masking parts of the words of the input sequence and attempting to restore them. This design allows BART to excel in generative tasks such as text summarization and machine translation. Wang et al.\cite{wang2022open} took the lead in decoding open domain words for the first time based on ZuCo data set, and the vocabulary size expanded from about 250 to about 50000. The team of HIT\cite{feng2023aligning} took into account the gap between topic and semantic on the basis of \cite{wang2022open}, and then retrained the EEG data to narrow the gap between topics. The semantic features of the data are more complete, and the SOTA performance at that time is achieved. The main innovation of DeWave\cite{duan2023dewave} is that it does not use eye movement dataset which believes that people do not necessarily read words in order, and even the segmentation using eye movement data is not the corresponding segmentation of data. Therefore, it is the first time to try to realize the direct translation of raw brainwave to text, although the translation effect is not as good as the combination of EEG data and eye-tracking data. But at least it proves that it works.

In this section, the machine learning algorithms employed in human-computer interaction tasks related to brain-inspired computing are segmented into two groups: feature extraction algorithms and classification algorithms. Within each category, we explore the relevant scenarios and limitations to provide a comprehensive understanding of their applicability.



  
 

\section{DEEP LEARNING MODELS FOR BRAIN-INSPIRED COMPUTING}
\label{section3}

This section will discuss in detail the recent and current contributions and research achievements of brain-inspired computing models for HCI tasks based on deep learning which are classified and listed in Table \ref{table2} according to the different brain signals processed. And we will elaborate the application of key technologies in the model.

\begin{table*}[htbp]
\footnotesize
  \centering
  \caption{Progress in deep learning models based brain-inspired computing tasks.}
  \resizebox{\linewidth}{!}{ 
    \begin{tabular}{ccr}
    \toprule
    \multicolumn{1}{c}{Task} & \multicolumn{1}{c}{Method Category} & \multicolumn{1}{l}{Representative Work} \\
    \midrule
        \multirow{28}[2]{*}{Model Based on Deep Learning} & \multirow{6}[1]{*}{EEG} &  \multicolumn{1}{l}{Automatic Speech Recognition, \cite{krishna2019state}} \\
          &       & \multicolumn{1}{l}{EEGNet, \cite{cooney2020evaluation}} \\
          &       & \multicolumn{1}{l}{EEG-transformer, \cite{lee2022eeg}} \\
          &       & \multicolumn{1}{l}{EEG-to-text, \cite{wang2022open}} \\
          &       & \multicolumn{1}{l}{CSCL, \cite{feng2023aligning}} \\
          &       & \multicolumn{1}{l}{DeWave, \cite{duan2023dewave}} \\
        \cmidrule{2-3} 
          & \multirow{8}[0]{*}{fMRI} & \multicolumn{1}{l}{Context Into Language,\cite{jain2018incorporating}} \\
          &       & \multicolumn{1}{l}{Brain2word, \cite{affolter2020brain2word}} \\  
          &       & \multicolumn{1}{l}{multi-timescale models, \cite{jain2020interpretable}} \\ 
          &       & \multicolumn{1}{l}{Semantic-based Classification, \cite{lin2022neural}} \\  
          &       & \multicolumn{1}{l}{Self-supervised Learning, \cite{millet2022toward}} \\  
          &       & \multicolumn{1}{l}{Cross-modal cloze task, \cite{zou2022cross}} \\   
          &       & \multicolumn{1}{l}{UniCoRN, \cite{xi2023unicorn}} \\   
          &       & \multicolumn{1}{l}{Semantic Reconstruction, \cite{tang2023semantic}} \\  
        \cmidrule{2-3} 
          & \multirow{5}[0]{*}{MEG} & \multicolumn{1}{l}{Seq2seq Learning,\cite{dash2019automatic}} \\   
          &       & \multicolumn{1}{l}{Decoding Speech(Dash D,et al.),\cite{dash2019decoding}} \\ 
          &       & \multicolumn{1}{l}{Decoding Imagined and Spoken Phrases, \cite{dash2020decoding}} \\  
          &       & \multicolumn{1}{l}{MEG Sensor Selection, \cite{dash2020meg}} \\  
          &       & \multicolumn{1}{l}{Decoding Speech(Défossez A, et al.), \cite{defossez2023decoding}} \\  
        \cmidrule{2-3} 
          & \multirow{5}[0]{*}{ECoG}& \multicolumn{1}{l}{Machine Translation, \cite{makin2020machine}} \\   
          &       & \multicolumn{1}{l}{Brain2char, \cite{sun2020brain2char}} \\   
          &       & \multicolumn{1}{l}{Neural Speech Decoding, \cite{chen2023neural}} \\   
          &       & \multicolumn{1}{l}{Direct Speech Reconstruction, \cite{berezutskaya2023direct}} \\   
          &       & \multicolumn{1}{l}{Synthesizing Speech, \cite{shigemi2023synthesizing}} \\    
    \bottomrule
    \end{tabular}%
    }
  \label{table2}%
\end{table*}%

\subsection{Data Acquisition And Preprocessing}

The acquisition of brain signal datasets often requires high cost and strict experimental conditions. Some high-quality public datasets provide great convenience for researchers in the field of brain-inspired computing. This section will focus on the construction of high-quality datasets from three aspects: data acquisition stage, data preprocessing stage and the public datasets.

\subsubsection{Data Acquisition}
In the latest research in the field of brain signals, several key techniques have been widely used. The quality of the data is guaranteed and improved, which is convenient for the relevant personnel of computer science to carry out subsequent algorithm research, and the following is a brief introduction to these key technologies:

\textbf{Stimuli \& experimental design:} 
In general, event-related potential(ERP) experiments use stimuli that are best perceived as simple and clear. If the properties of the stimulus (especially the content, intensity, duration, or location of the stimulus) are not relevant to the purpose of the experimental study, they need to be appropriately controlled. The so-called "control" is to be as consistent as possible between each spur within the condition and between different conditions. If a parameter of the stimulus is not possible to be consistent in practice, the researcher should also try to reduce the variability of the parameter within the condition, and the difference between the conditions should be statistically tested to ensure that the difference is not significant.

In the field of psychology, the event-related potential technology is often used to study, which is also a great inspiration for the design of brain signal acquisition experiments\cite{sur2009event}. Experimental design, especially the design of experimental stimuli, greatly affects the quality of collected data, and further affects the algorithms of data preprocessing and the modeling process of brain signal decoding. In order to improve the strength of brain signals, researchers often design experiments by combining internal stimuli and external stimuli. External stimuli such as visual, auditory, and tactile stimuli can further strengthen internal stimuli, especially for cognitive tasks of brain signal decoding and some emotional perception tasks. For example, when designing visual stimuli, attention should be paid to the center of visual materials (such as pictures) to reduce or even avoid the appearance of eye movement artifacts. If the visual materials are text stimuli, certain control problems can be considered to ensure the accuracy of brain signals and enhance the strength of brain signals. When designing auditory stimuli, the properties of the audio material itself such as pitch, loudness and discrimination acuity should be taken into account. When designing tactile stimuli, one can focus on the effects of the duration and frequency of the stimuli. Most of the public data sets have been verified by a large number of experiments, and strict experimental schemes are used to focus on the data set collection task itself, which can ensure the quality of the collected data, but it may not be suitable for all experiments. When the researchers need to design their own experiments, we hope that the above tips in terms of experimental design can bring some inspiration and help.

\textbf{Eye-tracking acquisition:} 
Eye-tracking serves as an indirect indicator of cognitive engagement, as gaze patterns exhibit a strong association with the cognitive workload at different stages of human text processing\cite{rayner1998eye}. For instance, fixation duration tends to be prolonged for lengthy, uncommon, and unfamiliar words\cite{just1980theory}. Each dataset encompasses distinct eye-tracking features, with the most prevalent features, namely first fixation duration, first pass duration, mean fixation duration, total fixation duration\cite{hollenstein2021leveraging}, and number of fixations, consistently available across all seven datasets\cite{hollenstein2019cognival}.

\textbf{EEG acquisition:} 
Electroencephalography (EEG) serves as a means to record the intricate electrical activity of the brain, capturing voltage fluctuations across the scalp with exceptional temporal precision. Hauk\cite{hauk2004effects} has presented compelling evidence supporting the modulation of early electrophysiological brain responses influenced by word frequency, highlighting the rapid nature of lexical access triggered by written word stimuli within a timeframe of less than 200 ms post-stimulus presentation. The EEG datasets employed in this study\cite{hollenstein2021leveraging} were derived from sessions involving either the reading of sentences or the listening to natural speech. The extraction of word-level brain activity was achieved through diverse methodologies, including mapping to eyetracking cues\cite{hollenstein2018zuco}, alignment with auditory triggers\cite{broderick2018electrophysiological}, capturing solely the terminal word in each sentence\cite{ettinger2016modeling}, or employing the method of serial word presentation\cite{frank2013reading}. Preprocessing of EEG data adhered to standardized procedures across all four data sources, encompassing essential steps such as band-pass filtering and artifact removal\cite{hollenstein2019cognival}.

\textbf{fMRI acquisition:}
Functional magnetic resonance imaging (fMRI) is a technique utilized to measure and map brain activity by detecting changes correlated with blood flow\cite{hollenstein2021leveraging}. With a temporal resolution of two seconds, fMRI captures alterations in blood flow, allowing one scan to encompass multiple words during continuous stimuli like natural reading or story listening. The datasets usually include both isolated stimuli\cite{mitchell2008predicting} and continuous stimuli\cite{wehbe2014aligning}. While it is comparatively simpler to extract word-level signals from isolated stimuli, continuous stimuli offer the advantage of extracting signals in context across a broader vocabulary. FMRI data consists of representations of neural activity within millimeter-sized cubes known as voxels. As part of the standard fMRI preprocessing procedures prior to the analysis, various preprocessing methods, including motion correction, slice timing correction, and co-registration\cite{hollenstein2021leveraging}, were implemented\cite{hollenstein2019cognival}.

In summary, the design of experimental stimuli has a profound impact on the quality of the data set, and certain norms should be followed, such as controlling irrelevant variables and combining internal and external stimuli to enhance the stimulus signal. Data collection is the key step. After the original data collection is completed, it also goes through a series of preprocessing such as denoising and data segmentation. Subsequent work will use the processed data to complete the model construction, encoding and decoding.

\subsubsection{Data Preprocessing}
Data preprocessing is crucial in brain signal decoding, because it helps to remove noise and artifacts, better highlight signal components related to specific brain activities, improve signal quality and reliability, make subsequent feature extraction and model training more accurate and effective, strong decoding performance and robustness. Next, some common pretreatment steps of brain signals such as filtering and artifact removal are summarized, and the effects of different pretreatment steps on different brain signals are explained.

\textbf{Filter processing:} 
Filter processing is the most basic step in brain signal preprocessing, which is used to remove unwanted frequency components. EEG and ECoG utilize filtering to remove unwanted low and high frequency signals, and high-pass filtering, typically at 0.1-1 Hz, is used to eliminate low-frequency drift, such as slow voltage fluctuations in time that can result from poor electrode contact or certain condition changes in the scalp\cite{khosla2020comparative}. Low-pass filtering, typically between 30-100 Hz, is used to remove high-frequency interference such as muscle activity while maintaining frequency signals associated with brain activity. In particular, band-stop filters (50Hz) are often used to eliminate frequency interference from power cords, which often comes from laboratory equipment or external electronics\cite{sharma2022analysis}. The filtering process in MEG is similar, using bandpass filtering to remove low and high frequency signals of no interest such as muscle electrical activity or external high-frequency interference\cite{ferrante2022flux}, while accurately filtering out power line frequency noise to ensure the accuracy of the magnetic field signal. fMRI uses high-pass filtering to remove very low-frequency drifts, chronic drifts that are often affected by physiological noise such as heart rate and breathing, as well as changes in device temperature\cite{kaplan2022filtering}. The purpose of high-pass filtering is to exclude non-signal-related low-frequency changes to improve the stability of time series data.

\textbf{Artifact removal:} 
Due to the interference of various physiological and non-physiological factors, the brain signals often contain artifacts. During EEG and MEG, the electrode signal is susceptible to interference from eye movement, heartbeat, and muscle activity. Independent component analysis (ICA) is used to decompose and identify the independent signal sources, and then eliminate the components associated with artifacts to purify the signal. EEG artifact correction may also include manual examination and labeling of abnormal signal segments\cite{bullock2021artifact}. ECoG requires artifact removal, especially in relation to implant surgery. Artifacts of electrode displacement or non-neural activity that may occur during surgery, such as patient body movement, can be identified and corrected by simultaneous video recording to ensure the recorded signal purity\cite{islam2021signal}. Artifact correction in fMRI focuses on physiological fluctuations, such as those caused by heartbeat and breathing\cite{bullock2021artifact}. This is usually done by recording synchronized physiological signals and applying regression analysis methods to isolate the effects of these signals from the fMRI data with the goal of retrieving pure BOLD responses.

\textbf{Reference reset:} 
Reference reset of EEG and ECoG is necessary because these techniques rely on measuring the potential difference relative to a reference point. Using an average reference, that is, subtracting the average of all electrodes from the individual electrode signals, can reduce the non-uniformity of local areas, thus making the spatial distribution of signals more uniform and reliable. Other common reference methods include binaural reference, the use of symmetrical body surface electrodes, which help reduce the artifact effect of the overall reference position\cite{taghaddossi2024electroencephalogram}.

\textbf{Spatial correction:}
Avoiding head movement artifacts is crucial in fMRI, with the help of voxel alignment techniques such as rigid body transformation to correct head movements during acquisition\cite{krentz2023comparison}. Usually, the 3D rigid body alignment technique is used to adjust the geometric position of each image relative to the reference image, so as to avoid the error introduced by motion. In addition, slight head movements during MEG recording can affect results, so head positioning sensors are used to monitor and correct head movements in real time. ECoG Because of the physiological fixation of the electrode on the surface, the spatial correction is more related to post-operative fretting, and the electrode position is often recalibrated by reference markers\cite{duraivel2023high}.

\textbf{Time-frequency analysis:}
EEG and ECoG make extensive use of time-frequency analysis to understand the dynamic spectral characteristics of signals. Short-time Fourier transform (STFT) and wavelet transform are important tools to obtain frequency activity in different time periods\cite{jiao2024eeg}. Through these techniques, dynamic EEG rhythms such as $\alpha$ wave and $\beta$ wave can be analyzed in specific time periods. MEG data can also be interpreted by time spectrum analysis, and the inverse problem solving method can be used for further neural source localization analysis\cite{minarik2023optimal}. In fMRI, this type of analysis is used less, because fMRI focuses on spatial analysis and long-term changes in time series signals\cite{belhaouari2023pft}.

With these processing steps, researchers can more effectively remove interference, improve data quality, and ultimately obtain more accurate analysis of brain function and behavior. These steps work together to ensure that the most useful and accurate information is extracted in complex brain function studies.

\subsubsection{Public Dataset}
Before establishing the model, it is very important to select high-quality dataset. Although high-quality dataset can be obtained after a series of data preprocessing, due to the particularity of brain signals, data acquisition equipment and cost requirements are high, so the collection conditions are not necessarily available. Most experiments are completed based on existing public data sets. The most commonly used eye tracking datasets, EEG datasets, and fMRI datasets are summarized as shown in Table \ref{table3}.

\begin{table*}[htbp]
    \footnotesize
    \centering
    \caption{Cognitive data sources utilized in this study include the Coverage metric, representing the percentage of vocabulary in the data source that aligns with the list of most frequently occurring English words in the British National Corpus. The statistical results in the table are from \cite{hollenstein2019cognival}.}
    \resizebox{\linewidth}{!}{
        \begin{tabular}{ccccccc}
        \toprule
        \multicolumn{1}{c}{Data type} & \multicolumn{1}{c}{Data source} & \multicolumn{1}{c}{stimulus} & \multicolumn{1}{c}{subject} & \multicolumn{1}{c}{tokens} & \multicolumn{1}{c}{types} & \multicolumn{1}{c}{coverage} \\
        \midrule  
        \multirow{7}{*}{EYE-TRACKING} 
        & GECO(\cite{cop2017presenting}) & text & 14 & 68606& 5383 & 95\% \\
        & DUNDEE(\cite{kennedy2003dundee}) & text& 10 & 58598 & 9131 & 94\% \\
        & CFILT-SARCASM(\cite{mishra2016predicting}) & text& 5 & 23466 & 4237 & 85\% \\
        & ZuCo(\cite{hollenstein2018zuco}) & text& 12 & 13717 & 4384 & 90\%\\
        & CFILT-SCANPATH(\cite{mishra2018scanpath}) & text& 5 & 3677 & 1314 & 89\% \\
        & PROVO(\cite{luke2018provo}) & text& 84 & 2743 & 1192 & 95\% \\
        & UCL(\cite{frank2013reading}) & text& 43 & 1886 & 711 & 98\% \\
        \midrule 
        \multirow{4}{*}{EEG} 
        & ZuCo(\cite{hollenstein2018zuco}) & text & 12 & 13717& 4384 & 90\% \\
        & NATURAL SPEECH(\cite{broderick2018electrophysiological}) & speech & 19& 12000 & 1625 & 98\% \\
        & UCL(\cite{frank2015erp}) & text & 24& 1931& 711 & 98\% \\
        & N400(\cite{broderick2018electrophysiological}) & text & 9 & 150& 140 & 100\% \\
        \midrule 
        \multirow{4}{*}{fMRI}
        & HARRY PORTER(\cite{wehbe2014aligning}) & text& 8 & 5169 & 1295 & 92\% \\
        & ALICE(\cite{brennan2016abstract}) & speech& 27 & 2066 & 588 & 99\% \\
        & PEREIRA(\cite{pereira2018toward}) & text/image& 15 & 180 & 180 & 99\% \\
        & NOUNS(\cite{mitchell2008predicting})& image & 9& 60 & 60 & 100\% \\
        \bottomrule
        \end{tabular}%
    }
    \label{table3}%
\end{table*}

Eye-tracking technology enables researchers to capture participants' eye movements during silent reading with minimal guidance or interference from the researchers. Additionally, in contrast to lexical decision tasks, eye-tracking technology can capture linguistic representations as they naturally occur in everyday life, devoid of interference from additional decision components or response mechanisms typically present in lexical decision-making tasks. Utilizing state-of-the-art eye-tracking devices, the position of the eye can be precisely determined every millisecond with high spatial accuracy, generating a rich and detailed dataset. The recorded eye movements during reading are frequently employed to explore visual word recognition in context\cite{rayner1998eye,rayner200935th}. For example,GECO\cite{cop2017presenting} presents the first natural reading eye tracking corpus specifically for bilingual sentence reading, selecting participants based on their language history and collecting detailed measures of proficiency. These data are well suited for research at one or more language processing levels\cite{cop2017presenting}. In brain-inspired computing tasks, eye-tracking data can help determine word boundaries\cite{wang2022open}.

ZuCo\cite{hollenstein2018zuco} is a dataset that integrates EEG and eye-tracking recordings of individuals engaged in reading natural sentences. Comprising high-density EEG and eye-tracking data, ZuCo includes records from 12 healthy native English-speaking adults who spent four to six hours each reading natural English text. This dataset encompasses two standard reading tasks and one task-specific reading task, providing EEG and eye-tracking data for a total of 21,629 words across 1,107 sentences and 154,173 fixation points. The EEG and eye-tracking signals from this dataset serve as valuable resources for training enhanced machine learning models across various tasks, with a particular focus on information extraction tasks such as entity and relationship extraction, as well as sentiment analysis. Furthermore, the dataset proves instrumental in advancing research on human reading and language comprehension processes, delving into the intricate interplay between brain activity and eye movement at a granular level\cite{hollenstein2021leveraging}.

ZuCo2.0\cite{hollenstein2019zuco} introduces a novel and freely accessible corpus, capturing both eye-tracking and brain electrical activity during natural reading and annotation. Distinguishing itself as the inaugural dataset facilitating a direct comparison of these two reading paradigms, ZuCo2.0 meticulously details the material and experimental design, undergoing thorough validation to ensure the data's quality. Tailored for cognitively inspired Natural Language Processing (NLP), this corpus boasts broad potential for application and reuse. The furnished word level and sentence level eye-tracking and EEG features offer utility in enhancing and assessing NLP and machine learning methods. Moreover, given the inclusion of semantic relational labels and participant comments at the sentence level, the dataset proves valuable for relational extraction and classification endeavors. As an upgraded version of the ZuCo dataset. ZuCo2.0 takes into account the effect of time periods on the test results, intentionally repeating some stimulus text for both normal reading tasks and special reading tasks.

\begin{table*}
  \centering
  \caption{Illustration of three task data examples for the ZuCo dataset adapted from\cite{hollenstein2018zuco}.}
  \label{tab:table_0}
  \begin{threeparttable}
    \begin{tabular}{|>{\columncolor{gray!30}\centering\arraybackslash}m{2.3cm}|>{\centering\arraybackslash}m{4.2cm}|>{\centering\arraybackslash}m{4cm}|>{\centering\arraybackslash}m{4cm}|}
      \hline
      \rowcolor{gray!30} \textbf{Task} & \textbf{Emotion Classification} & \textbf{Entity Extraction} & \textbf{Relation Extraction}  \\
      \hline
      Material & Positive, negative, or neutral \newline sentences from movie reviews
      & Wikipedia sentences containing specific relations
      & Wikipedia sentences containing specific relations
      \\
      \hline
      Data Sample & The film often achieves a mesmerizing poetry (positive)
      & Talia Shire (born April 25, 1946) is an American actress of Italian descent.\newline
      Relations: nationality, job title
      & Lincoln was the first Republican president.
      Relation: political affiliation
      \\
      \hline
      Task Description & \makecell{Based on the previous sentence,\\ how would you rate\\ this movie from 1-5?
      }
      & \makecell{Talia Shire was a …\\1)singer 2)actress 3)doctor
      }
      & \makecell{Does the sentence contain the\\ political affiliation relation?\\
        1)Yes 2)No
      } \\
      \hline
    \end{tabular}
  \end{threeparttable}
\end{table*}

As shown in Table \ref{tab:table_0}, the three tasks and stimulus control conditions set in the open dataset ZuCo are listed. Subjects record EEG data by reading sentences and answering control questions, which provides high quality EEG datasets related to natural language processing tasks such as sentiment analysis and information extraction for computer science researchers.

Functional magnetic resonance imaging (fMRI) stands out among non-invasive neuroimaging techniques, offering the highest spatial resolution. This series of investigations commenced with Mitchell et al. in 2008, marking the inaugural demonstration of the feasibility of decoding words from fMRI data. The pioneering approach involved leveraging semantic representations of words and learning cross-modal mappings between fMRI images and word vectors\cite{mitchell2008predicting}. Mitchell et al. evaluated their learned neurodecoder through a pairwise classification task, a binary classification exercise distinguishing which of two stimuli corresponds to an fMRI image. Subsequently, this paired classification methodology has become widely adopted by researchers in non-invasive neural decoding for word decoding tasks\cite{zou2022cross}.

In summary, both EEG and fMRI are used to study brain activity, and are the two most commonly used brain signals to decode text or speech. Eye movement data sets can play an auxiliary role in decoding. These public datasets promote the research of decoding using deep learning techniques.


\subsection{EEG-to-text(speech)}
\begin{figure*}[htbp]
    \centering
    \includegraphics[width=0.8\textwidth]{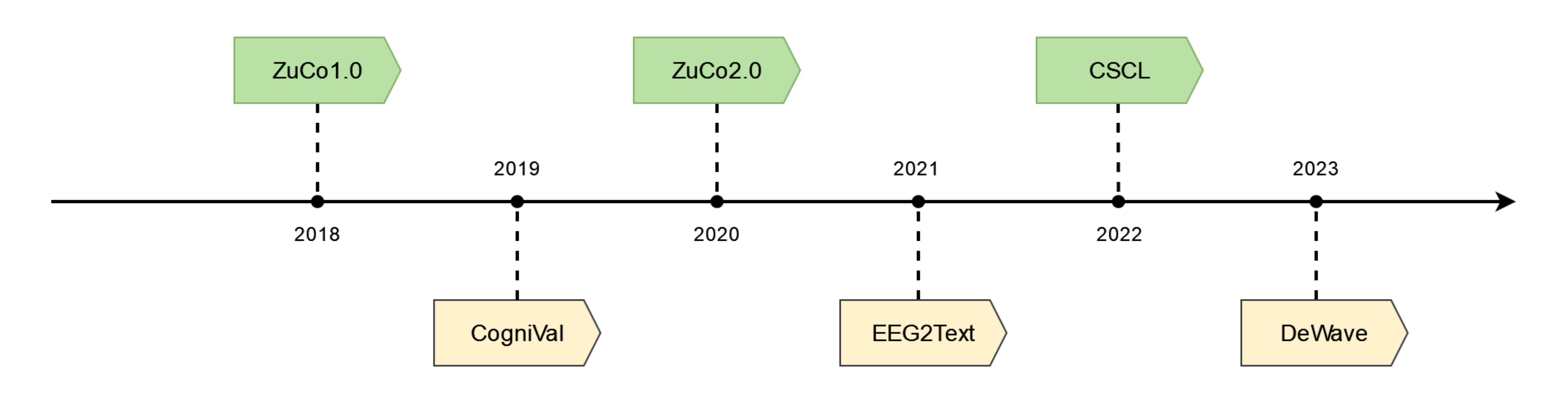}
    \caption{The development of brain-inspired computing models based on deep learning pre-trained language models with EEG signals in recent years. }
    \label{fig:2}
\end{figure*}

\begin{figure*}[htbp]
    \centering
    \includegraphics[width=0.86\textwidth]{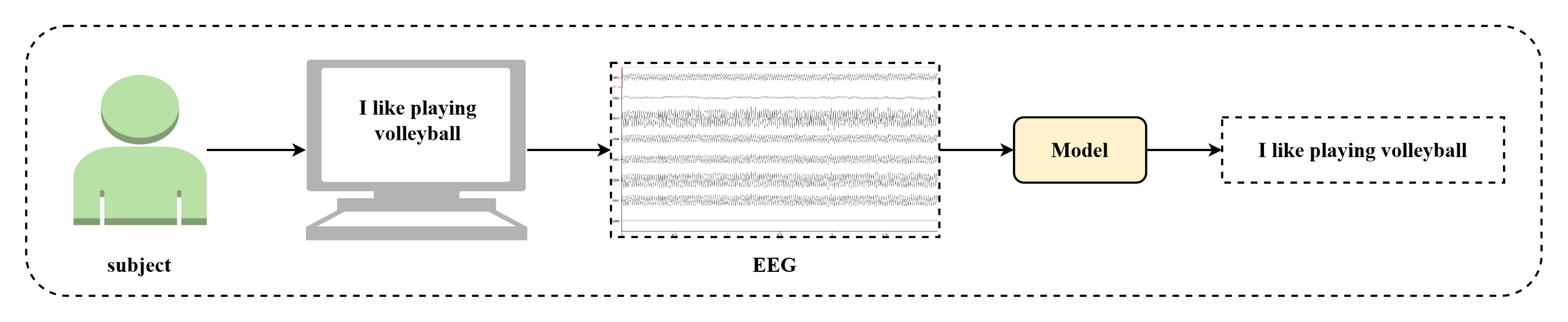}
    \caption{Illustration of the EEG-to-text generation task. The left portion depicts the EEG recording process, where a subject reads a sentence on the screen while EEG signals are recorded. Simultaneously, an eye-tracking device facilitates the precise definition of word boundaries through fixations. The objective of the task is to generate the sentence that elicits the recorded EEG signals(adapted\cite{feng2023aligning}).}
    \label{fig:3}
\end{figure*}

Sensory, language, emotion and other processes are very rapid, and the high temporal resolution of EEG makes it very suitable for capturing these fast, dynamic and sequential cognitive events. Therefore, many researchers are engaged in the study of decoding text or speech by EEG. The EEG-to-text (speech) task provides a convenient and widely applicable means of human-computer interaction by non-invasively measuring EEG electrical signals and converting them into text or speech. The current EEG-to-text tasks aim to assume that the human brain acts as a special text encoder, implementing an open vocabulary brain-to-text system using a pre-trained language model. Figure \ref{fig:2} demonstrates the development of brain-inspired computing models based on deep learning pre-trained language models with EEG signals in recent years. Figure \ref{fig:3} shows the EEG-to-text task flow diagram.The subjects read sentences on the screen, the researchers recorded the EEG signal during the reading process, and then decoded the text through the EEG signal.

\subsubsection{Recent Progress}
Decoding brain states into understandable representations has long been the focus of research\cite{kobler2022spd}. EEG signals are particularly favored by researchers because they are non-invasive and easy to record\cite{pan2022matt,wagh2022evaluating}. Traditional EEG decoding techniques mainly focus on classification tasks, such as emotion recognition\cite{bos2006eeg,jenke2014feature,song2018eeg,suhaimi2020eeg,wang2021review}, motor imagination\cite{saa2013discriminative,al2021deep,altaheri2023deep}, robot control\cite{millan2004noninvasive,tonin2011brain,gandhi2014eeg,bi2013eeg,salazar2017correcting}, and games\cite{nijholt2008bci,de2010defecting,liao2012gaming,kerous2018eeg,vasiljevic2020brain}. However, these task-bound tags are not sufficient for extensive brain-computer communication. As a result, interest in brain-to-text (speech) translation has surged in recent years.

For example, the researchers\cite{wang2022open} extend the problem to open word EEG on natural reading tasks to text sequence-to-sequence decoding and zero-sample sentence emotion classification. Assuming that the human brain acts as a unique text encoder, the authors introduce a novel framework that leverages pre-trained language models for the first time. The model achieved a BLEU-1 score of 40.1\% in EEG-to-text decoding and an F1 score of 55.6\% in zero-sample EEG-based ternary emotion classification, significantly better than the supervised baseline. Furthermore, the model demonstrates the capability to process data from diverse topics and sources, indicating substantial potential for high-performance open-vocabulary brain-to-text systems as more data becomes accessible.

In addition, the notable difference between subject-relevant EEG representations and semantically relevant text representations presents a significant challenge to this task. To address this challenge, Feng et al.\cite{feng2023aligning} propose a curricular Semantic Perception Contrastive Learning strategy (CSCL). This approach effectively recalibrates subject-dependent EEG representations into semantically dependent EEG representations, reducing variance. CSCL specifically groups together semantically similar EEG representations while separating distinct EEG representations. Furthermore, to introduce more meaningful contrast pairs, the researchers meticulously utilized curriculum learning, not only for generating significant contrast pairs but also for ensuring progressive learning. Subsequent research not only demonstrated its superiority in single-agent and low-resource settings but also showcased strong generalization in zero-sample settings.

Finally, with the rapid development of large language models such as ChatGPT\cite{bubeck2023sparks,floridi2020gpt}, in-depth research into ways to bridge the gap between brain signals and natural language representations becomes critical. However, the early work\cite{anumanchipalli2019speech,herff2015brain,makin2020machine,sun2019towards} and several models mentioned above, relied on external event markers like handwriting or eye-tracking fixations which may not be readily available. And the order of eye-tracking data may not correspond to the order of spoken words. To address these problems, the authors introduce a novel framework, DeWave\cite{duan2023dewave}, which integrates discrete coded sequences into the task of translating open vocabulary EEG to text. DeWave uses quantized variational encoders to derive discrete codex codes and align them with pre-trained language models. This discrete code representation mitigates sequence mismatches between eye fixation and spoken language and minimizes interference caused by individual differences in brainwaves by introducing text-EEG contrast alignment training which is the first to decode EEG signals first without the need for word-level sequential labeling scoring 20.5 BLEU-1 and 29.5 Rouge-1 on the ZuCo dataset.

To sum up, from decoding words to decoding text to decoding directly without the aid of eye-tracking data, from closed-domain words to open-domain words, researchers have used the latest deep learning techniques and methods to complete increasingly complex EEG-to-text decoding tasks.

\subsubsection{Key Technologies}
EEG-to-text models based on deep learning have experienced different stages from closed vocabulary to open vocabulary, from dividing word boundaries by external means to directly translating brain waves, and each model has its own contributions and key techniques.

\textbf{EEG-to-text Decoding:} 
The author tries to maximize the probability of decoding the sentence as the following equation:
\begin{equation}
    p(S|{\varepsilon})=\prod_{t=1}^{T}{p({s_t}\in{\nu}|{\varepsilon,s<t})}
\end{equation}
Where $T$ represents the length of the target text sequence. The primary challenge in the environment is that the vocabulary size, denoted as $|\varepsilon|$ (50,000), is considerably larger than in previous sequence-to-sequence studies (250)\cite{makin2020machine}.

\textbf{Curriculum Learning:} 
The researchers proposed an approach based on curriculum learning. The overall training process adopts a two-step approach. Firstly, CSCL is employed to train the pre-encoder. Formally, a contrastive triple $(E_i, E_i^+, E_i^-)$ is obtained for a given anchor $E_i$. After the transformation by the pre-encoder, the corresponding vectors become $(h_i, h_i^+, h_i^-)$, where $h_i$ is the averaged vector of the pre-encoder outputs. Following the contrastive framework in\cite{gao2021simcse,wang2023large,saeidi2021neural}, with $N$ as the mini-batch size, the objective is to minimize the cross-entropy loss $l_i$ defined by:
\begin{equation}
    l_i=-log{\frac{e^{{sim(h_i,h_i^+)}/r}}{{\sum_{j=1}^{N}}({e^{{sim(h_i,h_i^+)}/\tau}}+{e^{{sim(h_i,h_i^-)}/\tau}})}}
\end{equation}
where $\tau$ is a temperature hyperparameter. $sim(h_i,h_j)$ is the cosine similarity.

Secondly, leveraging the contrastive-trained pre-encoder, jointly fine-tune all parameters of the BRAINTRANSLATOR to minimize the cross-entropy loss in a parallel training corpus $(E, S)$\cite{wang2023large}:
\begin{equation}
    L = -\sum_{\substack{(E,S)\in(E,S)}} \log p(S|E; \theta)
\end{equation}

\begin{figure*}[htbp]
    \centering
    \includegraphics[width=0.86\textwidth]{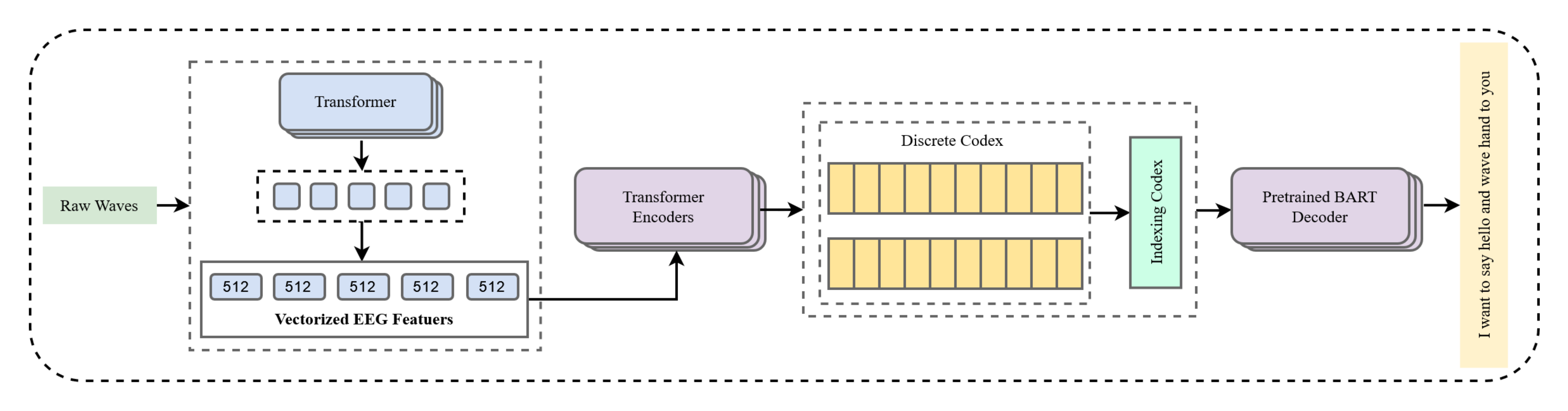}
    \caption{The DeWave model vectorizes the original EEG waves into embeddings, the vectorized features are encoded into latent variables by codex indexing into discrete latent variables, and finally, the pre-trained BART model converts this discrete codex representation into text(adapted\cite{duan2023dewave}).}
    \label{fig:4}
\end{figure*}

\textbf{Discrete Codex:} 
The key steps of the DeWave model are shown in Figure \ref{fig:4}. First, the original brainwave is vectorized, the vectorized features are encoded into hidden wave variables and converted into discrete latent variables by codex index. Finally, the pre-trained BART model converts this discrete codex into text.

Given the EEG waves, it is firsted vectorized into embedding, where $X$ is the embedding sequence as shown in Equation 4.  
\begin{equation}
    z_q(X)={z_q(x)}_i
\end{equation}
A codex book ${c_i}\in{R^{k*m}}$ is initialized with number $k$ of latent embedding with size $m$ as shown in Equation 5.
\begin{equation}
    z_q(x)=c_k
\end{equation}
The vectorized feature $X$ is encoded into $z_c(X)$ through a transformer encoder as shown in Equation 6. The discrete representation is acquired by calculating the nearest embedding in the codex of input embedding $x\in{X}$ as shown in Equation 4.
\begin{equation}
    k={argmin}_j||z_c(x)-c_j||_2
\end{equation}
Dewave directly decodes the translation output given the representation ${z_q}(X)$. Given a pre-trained language model, the decoder predicts text output.

In summary, with the help of deep learning technology, the study of EEG-to-text decoding task relies less and less on datasets, and the decoding accuracy is gradually improved

\subsection{fMRI-to-text(speech)}
As far as verbal stimuli are concerned, recent research has shown that fMRI scans can be decoded into embeddings of the words the subject is reading\cite{affolter2020brain2word}. The fMRI-to-text (speech) task uses functional magnetic resonance imaging for text conversion, and is able to capture deep brain activity due to its high spatial resolution, providing broad possibilities for human-computer interaction in pursuit of detailed and complex decoding. The research of decoding of text or speech based on fMRI is also gradually increasing and fMRI signals have been used more recently to decode text.

\subsubsection{Recent Progress}
FMRI provides the best spatial resolution of all non-invasive neuroimaging techniques\cite{zou2022cross}. Most of the fMRI-to-text based tasks are pairwise classification tasks\cite{wang2020fine,sun2019towards,sun2020neural}, or direct classification tasks\cite{affolter2020brain2word,zou2022cross}.

For example,the authors\cite{affolter2020brain2word} propose a model for decoding fMRI scans into words, and verify the decoding performance and generalization performance of the model through two tasks of pair-classification and direct classification. The direct classification task is to classify the brain scan directly into one of the $v$ words in the vocabulary under consideration. Because scanning is mapped to word categories rather than to representations of words, the task is not subject to limitations associated with specific vector representations. In addition, when there is no data from the target subject at the time of training, the reserve-one method is used to train the model with data from $n$-1 subjects and test on the remaining subjects, repeating the process for each subject. Due to the impossibility of subject-specific preprocessing, the model requires strong generalization, and the lack of consistency in FMRI scans across subjects and even across recorded sessions is a big challenge.

Due to the inherent noise in brain recordings, prior research has simplified brain-to-word decoding into a binary classification task, aiming to distinguish words corresponding to brain signals from incorrect ones. However, this paired classification approach faces limitations for practical neurodecoder development for two reasons. Firstly, it necessitates enumerating all pair combinations in the test set, making it inefficient for predicting words in a large vocabulary\cite{zou2022cross}. Secondly, even with a perfectly paired decoder, the performance of direct classification is not guaranteed. In order to address these challenges and advance realistic neurodecoders, the authors introduce a novel cross-modal cloze (CMC) task. This task involves predicting target words encoded in neural images using context as a cue. The results demonstrate the feasibility of decoding a single word from neural activity in a large vocabulary\cite{zou2022cross,wang2022open}.

Another representative advancement is UniCoRN. The disadvantage of fMRI is its low temporal resolution, and previous fMRI signal decoding methods often relied on feature extraction for predefined regions of interest (ROI), failing to make full use of time series information. UniCoRN  constructs an efficient encoder through snapshot and sequence reconstruction, allowing the model to deeply analyze the time dependence and maximize the extraction of information from brain signals. Specifically, UniCoRN contains two key stages: brain signal reconstruction and brain signal decoding. The brain signal reconstruction stage is divided into snapshot reconstruction and sequence reconstruction. The internal features of each snapshot and the temporal relationship of the snapshot sequence are integrated by training the encoder. In the brain signal decoding stage, the representation of the brain signal is converted into natural language, each fMRI frame is treated as a word-level representation of the "language spoken by the human brain", and real human language is generated through a text decoder\cite{xi2023unicorn}.

Finally, some researchers have proposed using fMRI signals to reconstruct auditory stimuli that subjects hear or imagine\cite{tang2023semantic}. To overcome the low temporal resolution of fMRI, the decoder employs a non-end-to-end strategy of guessing candidate word sequences, assessing the likelihood that each candidate triggers the currently measured brain response, and selecting the best candidate for decoding. In the experiment, a coding model was trained for each subject to predict the semantic representation of text stimuli and the corresponding brain response. The beam search algorithm is used to generate candidate sequences word-by-word and preserve the most likely continuation when detecting new words in brain activity. Finally, the most likely continuation is preserved by scoring the coding model.

Overall, these studies summarize cutting-edge models and methods for decoding fMRI signals in the brain-inspired field, deepening the understanding of the relationship between brain activity and cognitive processes such as language and perception.

\subsubsection{Key Technologies}
The brain-inspired computing model based on fMRI-to-text has developed from the original binary classification model to the cross-modal task, which makes the direct classification task possible.

\textbf{Brain Decoding Model:}
In the model developed in this paper\cite{affolter2020brain2word}, the decoder of the model takes the one-dimensional vector of the fMRI scan as input, filling it as needed. By simply changing the output layer and the loss function, the model can be used for paired classification tasks. The loss is calculated as follows:
\begin{equation}
   \zeta_{\text{reg}} = \sum_{i=1}^{v} \cos(y_{\text{pr},i}, y_{\text{true},i}) - \sum_{j \neq i}^{v} \cos(y_{\text{pr},i}, y_{\text{true},j})
\end{equation}
Where $y_{\text{pr},i}$ is the predicted word embeddings of the word $i$, $y_{\text{true},j}$ is the true word embeddings of the word $j$, and $cos()$ is the cosine distance.

\textbf{Semantic Features Fusion:} 
Semantic features are extracted from fMRI images and can be seamlessly integrated directly into hidden states within the BERT embedding layer. To address the centrality problem\cite{radovanovic2010hubs}, an intermediate word embedding is introduced, and a retrieval-based method is devised. The fundamental concept involves utilizing the intermediate word embedding for cross-modal mapping, followed by transforming the prediction into the BERT embedding space through retrieval\cite{zou2022cross}.

The feature vector $f_i$ for fMRI test sample is computed as follows:
\begin{equation}
    f_i={\frac{1}{k}}\sum_{t=1}^{k}{w'}{j_t}
\end{equation}
Where $i = \{1, \dots, N\}$ and $w'$ denotes BERT embedding.

To be specific, let $hi_{mask}$ denote the hidden states of the [MASK] token, try to directly update $hi_{mask}$ using the following equation:
\begin{equation}
    h_{mask}^{i}=(1-\alpha){h_{mask}^{i}}+{\alpha{f_i}}
\end{equation}
where $\alpha\in{[0,1]}$ is a tuning parameter that controls how much information to fuse in.

\begin{figure}[htbp]
    \centering
    \includegraphics[width=0.45\textwidth]{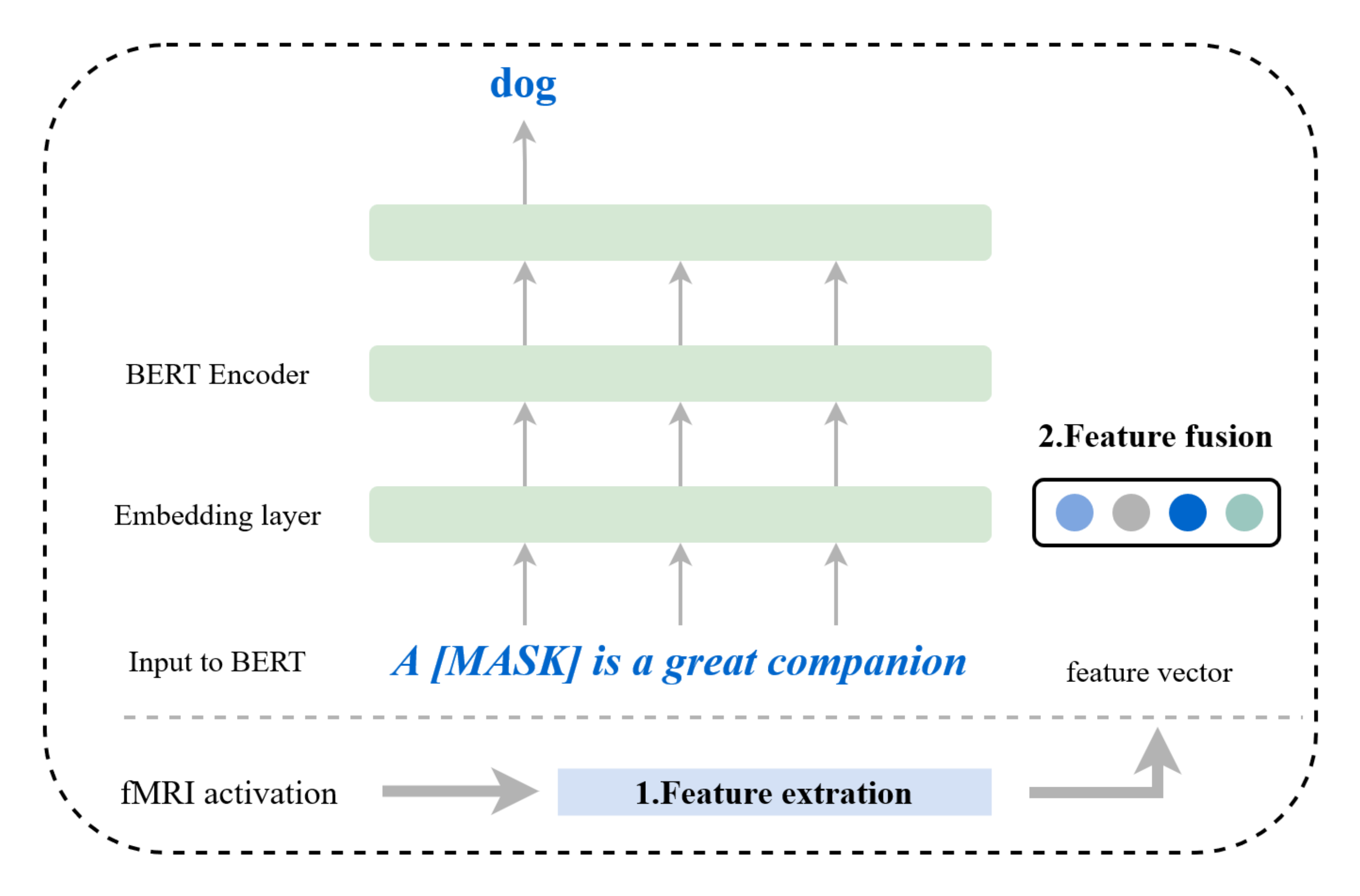}
    \caption{Illustration of the semantic feature fusion method. The pre-trained language model BERT is used to predict target words(adapted\cite{zou2022cross}).}
    \label{fig:5}
\end{figure}

As illustrated in Figure \ref{fig:5}, the objective is to incorporate feature vectors extracted from fMRI images into BERT for enhanced word prediction. The rationale is that if the feature vector contains valuable information about the target word, integrating it into the model should enhance prediction performance, going beyond using context alone as input\cite{zou2022cross}.

In summary, these key technologies have innovated the research method of fMRI-to-text task, improved the decoding accuracy, and provided new ideas for subsequent research.

\subsection{MEG-to-text(speech)}

MEG records magnetic field changes resulting from current flow in the brain\cite{cohen1972magnetoencephalography} and is capable of obtaining real-time resolution recordings of brain activity at a higher spatial resolution than EEG or fMRI\cite{proudfoot2014magnetoencephalography,mihelj2021machine}. According to our research, MEG signals have been used more recently to decode speech. The unique data properties of MEG make it suitable for brain disorders that are time- and space-sensitive. The MEG-to-text (speech) task decodes signals by monitoring magnetic activity in the brain, supporting human-computer interaction applications that require fast responses with its superior temporal resolution. Additionally, MEG is quiet and, therefore, user-friendly. Prior MEG studies on speech perception hold promising prospects for decoding speech production in brain activity signals\cite{suppes2000brain,chan2011decoding,guimaraes2007single,wang2017towards,mihelj2021machine}. 

\subsubsection{Recent Progress}
The convergence of advanced methodologies and technologies has paved the way for breakthroughs in understanding the spectral and temporal characteristics of MEG signals.

For example, the researchers delved into the spectral and temporal characteristics of MEG signals and trained these features using CNN to classify neural signals corresponding to specific phrases. Experiments show that CNN is remarkably effective in decoding speech in perception, imagination and generation tasks. In order to solve the problem of long CNN training time, the researchers used PCA algorithm to reduce the spatial dimension of MEG data, and carried out the model initialization through transfer learning. The experimental results show that the accuracy of speech decoding in speech generation stage is significantly higher than that in perception and imagination stage\cite{dash2019decoding}.

In addition, some researchers propose a single architecture data-driven approach for decoding natural language from MEG signals. Convolutional neural networks are used as encoders of brain signals and trained by contrast targets to align deep audio representations generated by the pre-trained speech self-supervised model wav2vec-2.0\cite{baevski2020wav2vec}. In addition, the contrast loss of the CLIP\cite{radford2021learning} model is used to align deep representations of the two modes of text and image, and the model can efficiently encode multiple levels of linguistic features and has a linear relationship with brain activation. The method and ideas provide a basis for subsequent work that can be effectively migrated to other tasks, including decoding continuous text\cite{defossez2023decoding}.

In summary, as a non-invasive method, MEG has garnered attention from research groups exploring its application in decoding speech or text. These advances represent the latest in the current MEG-to-text(speech) task.

\subsubsection{Key Technologies}
In the cutting-edge research of MEG signal decoding technology, the key technologies such as convolutional neural networks, transfer learning and signal alignment have not only deeply influenced the theoretical cognition of neuroscience, but also promoted the innovation in the field of speech decoding.

\textbf{CNNs and Transfer Learning:} 
This paper\cite{dash2019decoding} proposes an approach that the MEG signal was denoised by wavelet using machine learning technology, and then the PCA space dimension was reduced and the PCA coefficient spectrum was generated. A 3-layer 2D-CNN was then trained using deep learning techniques for each of the 200 gradient sensors. Conv3 features were then extracted from the 200 CNNs trained to further classify the phrases. Finally, transfer learning is used to reuse the pre-trained model for new data, and transfer the weight of the pre-trained network to the new network to obtain faster convergence speed and better performance. However, the researchers did not use these predefined networks in MEG signals, so the approach was to first perform neural speech decoding by training CNNS on a single subject's data, and then transfer the learning weight of that subject's trained CNNS to the next subject's speech decoding. The weights of the classification layer, softmax layer, and FC layer were not trained using data from another subject\cite{dash2019decoding}.

\begin{figure}[htbp]
    \centering
    \includegraphics[width=0.45\textwidth]{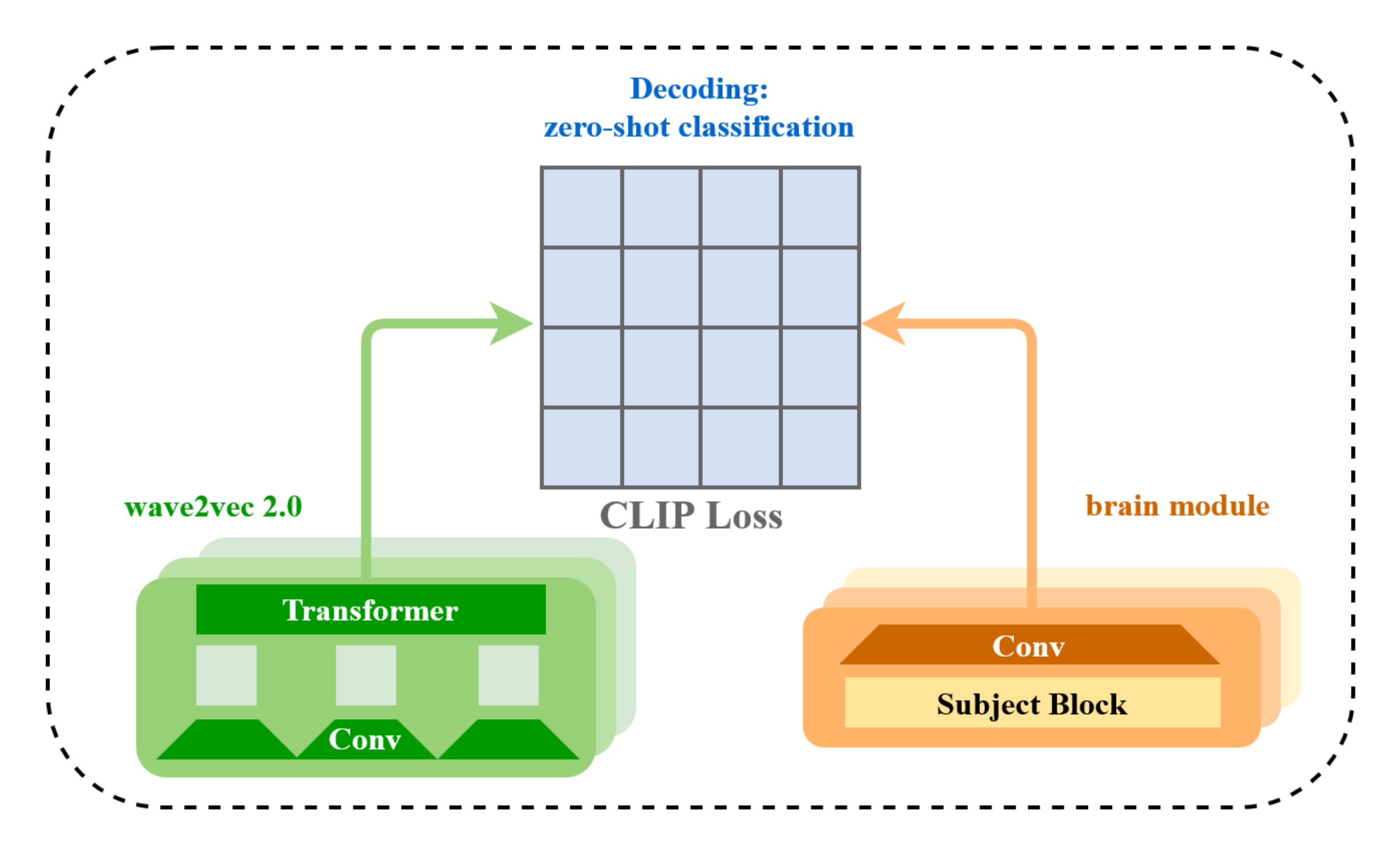}
    \caption{The model of the zero shot MEG-to-speech decoding method(adapted\cite{defossez2023decoding}).}
    \label{fig:6}
\end{figure}

\textbf{Signal Alignment:} 
As shown in Figure \ref{fig:6}, the researchers decoded speech from brain activity recorded using magnetoencephalograms or electroencephalograms in healthy participants as they listened to sentences. To do this, their model extracts deep context representations of 3s speech signals from the pre-trained speech module wav2vec2.0 and learns the brain's representations' activity on the corresponding 3s window to maximize alignment with these speech representations with contrast loss. Models with missing sentences were input at the time of assessment, and the probability of each 3-second speech fragment given each brain representation was calculated. Thus, the final decoding can be a "zero sample" because the audio snippet predicted by the model does not need to be present in the training set\cite{shigemi2023synthesizing}.

These technical advances highlight the continuous evolution of MEG signal decoding, and through the integration of deep learning models, speech processing models and neural signals, the technical gist of MEG decoding is deeply understood.

\subsection{ECoG-to-text(speech)}
ECoG is superior to surface EEG in terms of spatiotemporal resolution and signal-to-noise ratio\cite{shigemi2023synthesizing}. The ECoG-to-text(speech) task obtains high-resolution data by directly recording the electrical activity of the cortex, and while providing accurate and fast decoding, it is suitable for human-computer interaction scenarios that require highly accurate and rapid control. According to our research, ECoG signals have been used more recently to decode speech. It is particularly suitable for analyzing brain activity associated with speech in high gamma-band\cite{crone1998functional,pei2011spatiotemporal,shigemi2023synthesizing}.

\subsubsection{Recent Progress}
This section will focus on two recent studies that cover the exploration of decoding speech through electrical corticography (ECoG) signals and parsing continuous text using machine translation models.

For instance, employing a deep neural network-based encoder and a pre-trained neural vocoder, the researchers effectively reconstructed spoken sentences from ECoG signals\cite{shigemi2023synthesizing}. When 13 participants spoke short sentences, the ECoG signals were recorded, and the ECoG recordings could be mapped to spectrographs of spoken sentences using two-way Long short-term memory (BLSTM) or Transformer, comparing the performance of the two encoders. In addition, the combination of encoders and neural vocoders was evaluated, and the experiment found that the use of Transformer encoders for spectral graph prediction is better than that of BLSTM encoders, and their combination with neural vocoders can effectively synthesize speech\cite{shigemi2023synthesizing}.

In the past research on decoding brain activity, decoding of continuous text is relatively insufficient and ineffective. In order to deal with this problem, the researchers consider the brain signal as the source language and the corresponding continuous text as the target language, and use the machine translation model method to design a simple encoder-decoder structure neural network to decode the continuous text in the ECoG signal\cite{makin2020machine}. By introducing auxiliary loss in the training stage, the model is forced to accurately predict the audio representation of the speech at the corresponding time based on the hidden layer representation of the encoder, so as to improve the information capture ability of the encoder. The study achieved significant accuracy improvements, and the average word error rate for some participants was reduced to 7\%, providing a valuable reference for future research.

In summary, ECoG as an invasive data acquisition method has the highest spatial resolution but also has significant limitations, and these new developments further enhance the understanding of ECoG-to-speech decoding tasks.

\subsubsection{Key Technologies}
ECoG technology has high spatial and temporal resolution when recording electrical activity in the brain. Because ECoG is closer to the surface of the brain, it can provide more accurate and fine-grained information about neural activity. Some of the latest techniques for ECoG signal decoding also incorporate DL models.

\textbf{Decoding Approach:} 
The researchers conducted a comparison between the Transformer encoder and the BLSTM encoder. The Transformer encoder comprises $X$ identical layers, each consisting of two sub-layers: a multi-head self-attention mechanism and a fully connected feedforward network. When these two sub-layers are replaced by BLSTM, the model is referred to as a BLSTM encoder\cite{shigemi2023synthesizing}. Residual connection and layer normalization strategies are applied in each of the two sub-layers. The output from $X$ stacks of the same layer is fed into a second-layer feedforward network to represent a sequence of log-mel spectrographs. During the training of the encoder model, the output is utilized to approximate a sequence of log-mel spectrographs with a length $N$ equal to the output size of the time convolution layer\cite{shigemi2023synthesizing}.

\begin{figure*}[htbp]
    \centering
    \includegraphics[width=0.86\textwidth]{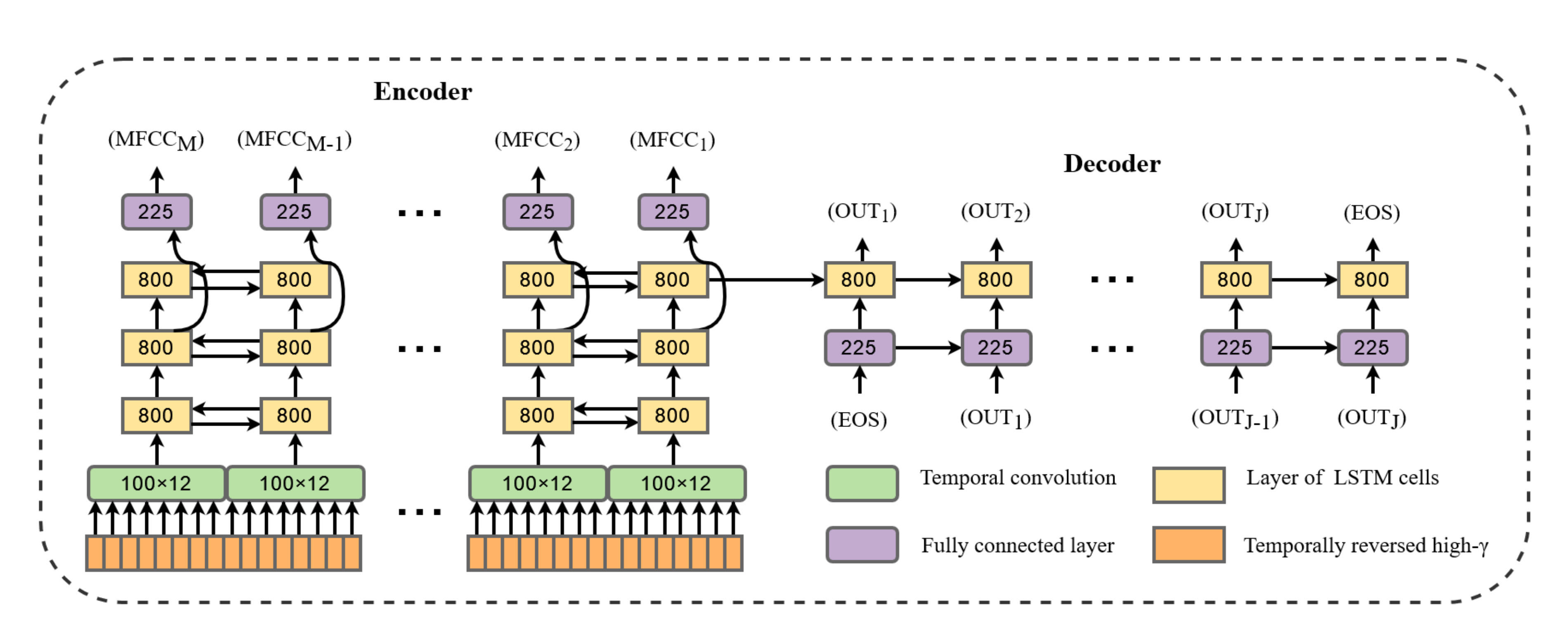}
    \caption{Illustration of the machine translation network architecture. The encoder and decoder are shown unrolled in time(adapted\cite{makin2020machine}).}
    \label{fig:7}
\end{figure*}

The model first uses time-span convolution for feature extraction, down-sampling the timing features to 16HZ and input the features into the LSTM network to generate continuous text. In order to guide the encoder to capture meaningful information, in addition to the end-to-end ECoG signal decoding text training, an additional auxiliary loss is introduced during the training phase. This loss requires the model to accurately predict the audio representation of the speech at the corresponding moment, based on the hidden layer representation of the encoder at each time step\cite{makin2020machine}. The network architecture is shown in Figure \ref{fig:7}, with encoders and decoders displayed in time expansion, i.e. sequence elements. Encoder and decoder all layers in the same row have the same input and output weights, and arrows in both directions represent bidirectional RNN.

\textbf{Neural Vocoder:} 
Parallel WaveGAN\cite{yamamoto2020parallel} is used to synthesize speech waveforms. Parallel WaveGAN is a waveform generator based on a non-autoregressive WaveNet model that is trained to synthesize speech based on a given Mayer spectrogram. The authors use an existing vocoder model pre-trained on a corpus dataset containing Japanese speech\cite{shigemi2023synthesizing}.

These two core technologies provide powerful ways to delve deeper into the relationship between brain signals and speech.

In this section, the tasks associated with brain-inspired computing for human-computer interaction using deep learning models are reviewed in detail from five perspectives. In addition to perspectives on four common brain signals, we highlight the importance of datasets.

\section{CHALLENGES AND SOLUTIONS}
\label{section4}
Decoding language from non-invasive brain activity is of increasing interest to researchers in neuroscience and natural language processing. Human beings have the ability to adapt to the complex environment, the ability to learn new things independently and the ability to cooperate with a variety of different cognitive abilities, but since the birth of artificial intelligence, no general intelligence system has reached the human level. The application of deep learning models in brain-inspired computing for HCI tasks faces several challenges and limitations. These issues must be seriously addressed and resolved to promote the development and application in this field. 
In this section, divided into six parts from three levels: model training, model application, and ethical issues, we will delve into the challenges faced by brain-inspired computing models and propose potential solutions. Specifically, challenges related to model training are elaborated in sections 4.1, 4.2, and 4.3, challenges related to model application are discussed in sections 4.4 and 4.5, and finally, ethical issues are analyzed in section 4.6.
The following are the factors to analyze the challenges and future research direction of these tasks:

\subsection{Challenges for Brain Dataset}
Most high-performance methods require data from invasive devices, such as ECoG, which require extremely high conditions to be collected and studied\cite{sliwowski2022deep,sliwowski2023impact,sergeev2023simple}. Non-invasive device data acquisition, such as fMRI, also has a large equipment threshold, and only professional institutions have the acquisition conditions. In addition, due to the limitation of word-level features, many current models based on EEG data sets need to use eye movement datasets and other external means to demarcate word boundaries, which indirectly increases the difficulty and cost of data collection. As a result, there are few public high-quality data sets. For example, most of the EEG-based brain signal decoding tasks are based on the public ZuCo\cite{hollenstein2018zuco} and ZuCo2.0\cite{hollenstein2019zuco}. The range of vocabularies is also limited, and many models are based on closed vocabularies, which do not fully conform to the concept of "silent speech" in direct thought translation by human brain. Deep learning models typically require large datasets, and their lack of generalization and robustness in small sample environments has not been fully addressed.

Although the data obtained by invasive devices is more accurate, researchers should be encouraged to use data sets collected by non-invasive devices to improve model performance by improving technical means such as word embedding methods. For example, achieving direct EEG-to-text translation is a great idea, and DeWave\cite{duan2023dewave}'s introduction of discrete coding similar to word embeddings could be a good step toward closing the gap between brainwaves and high-level language models. Expand the scope of the vocabulary and conduct research on the open vocabulary to improve the usefulness of future brain-computer interface.

\subsection{Challenges for Computational Resource Demand}

When deep learning models are applied to the training and decoding of brain signal data, the models often have huge demands on huge data sets and high computational resources. High-performance deep learning models often require extensive training data to accurately capture complex patterns and features, but in brain signaling research, large-scale data sets are difficult to obtain because the data acquisition process is complex and limited by equipment conditions and ethical considerations. In addition, these models are less robust in small sample environments, which means that even though training costs are high, the models struggle to perform well when the data is insufficient. In practical applications, due to the limitation of hardware resources, such as the processing power of smart phones or the computing power of portable devices, this high demand for computing resources often becomes an important bottleneck in the wide application of deep learning models.


To address these challenges, researchers can adopt a variety of strategies to optimize the training and reasoning processes of deep learning models. First, by introducing transfer learning techniques, using models that have been trained on similar tasks can significantly reduce the need for large-scale data sets and improve model performance in small sample cases. Second, through data enhancement techniques, small data sets can be manually expanded to improve the generalization ability of the model on more diverse data. In addition, exploring unsupervised learning methods can reduce the dependence on data annotation and discover useful information in data structures. In terms of computing resources, the development of more efficient training and inference algorithms, as well as the deployment of models on specialized hardware that supports fast parallel computing, can reduce resource requirements without significantly sacrificing performance. Combined, these methods can help researchers effectively apply deep learning models in resource-constrained environments, thereby expanding their feasibility and utility in real-world applications.

\subsection{Accuracy And Generalization of The Model}
The most advanced brain-to-text systems have achieved a degree of precision in using neural networks to decode language directly from brain signals. However, many current models are limited to small closed vocabularies, which are far from sufficient for natural communication. Although DeWave\cite{duan2023dewave} enhances EEG to text translation using discrete codex and raw wave coding, its accuracy is still poor compared to traditional language-to-language translation. In addition, due to the limitations of data collection in EEG-to-text tasks, the maximum number of subjects in the open dataset is 18 at one time\cite{hollenstein2019zuco}. Moreover, due to the particularity of EEG data, there are great differences among different subjects, so the improvement of model generalization performance is a long-term research direction.

For example. for the fMRI-to-text tasks, how to align fMRI signals with individual words when the stimulus is presented as a continuous time series of words is the basis for translating brain activity into text\cite{zou2022cross}. For the EEG-to-text tasks, we can start from the perspective of word embedding algorithm, compare the characteristics of different word embedding algorithms\cite{hollenstein2019cognival}, and improve the accuracy and generalization performance of the model by taking advantage of generative model and machine translation model. We can also start from the perspective of data sets, because different languages also have a great impact on the generalization performance of the model of converting EEG data into text. The future needs to collect larger EEG text data sets and extend the current framework to multilingual environments.

\subsection{Challenges for Real-time Processing}
In human-computer interactive applications, brain-inspired computing models need to deal with the challenges of immediacy and diversity brought by user brain signal input when combined with deep learning. First, in order to provide a smooth user experience, the model must respond with extremely low latency times, which puts higher real-time requirements on complex deep learning models. Secondly, human-computer interaction scenarios often require the model to process multi-modal and multi-channel data, such as the combination of EEG data and ophthalmic data, which increases the computational complexity and the burden of data processing. In addition, these interactions often take place in resource-constrained environments, such as smartphones or wearables, which have limited computing power and memory, making it more difficult to deploy and operate models efficiently.

Potential solutions to improve the real-time performance of brain-inspired computational models include a variety of strategies. First, model parameters can be reduced by using model compression and optimization techniques such as knowledge distillation and pruning, effectively reducing the computational burden and speeding up response times. Secondly, deploying edge computing can carry out part of the data processing task on the terminal device, reducing the delay of data transmission and achieving more immediate feedback. Using dedicated hardware such as Gpus in mobile devices or dedicated neural processing units can improve the speed and efficiency of model execution. In addition, lightweight and efficient network architecture is designed, and asynchronous processing and pipeline optimization allow multiple data streams to be processed simultaneously. This not only ensures efficient operation in multimodal interactions, but also ensures persistent real-time responses, thereby enhancing the overall human-computer interaction experience.


\subsection{Challenges for BCI Technology}
The current development of human-computer interaction based on brain-inspired computing is still far from achieving the performance expected by human beings, and faces many challenges: for example, a significant feature of brain signals is high-dimensional and multi-modal, how to obtain more brain data is a difficulty, and the correspondence between brain signals and motion, emotion, language, etc., needs to be explored more.

In addition, the common non-invasive EEG signals have several problems: low signal-to-noise ratio, low reliability, and unstable signals. These challenges lead to the need for further exploration and research on efficient algorithms to improve accuracy, solve the curse of dimensionality, and reduce the time cost.

Moreover, BCI systems require users to be highly concentrated during training, which will lead to high workload of brain-computer interaction such as visual fatigue. Meanwhile, the volume and shape of brain-computer interface devices need to be further miniaturized to make users more comfortable and portable.

Furthermore, the broadening of brain channels, the implementation and performance improvement of brain-computer interface technology give humans the ability to directly interpret and control thoughts, which inevitably brings new challenges to human privacy rights, and there is also the risk of hacking, which touches the ethical bottom line.

\subsection{Challenges for Ethical Concerns}
The ethical problems of brain-inspired computing models mainly focus on several aspects. First, the privacy protection of physiological signal data is a key issue, because these models need to process a large amount of sensitive brain signal data, which can lead to personal privacy disclosure if improperly used or leaked. Second, algorithmic bias is also an important ethical consideration. When facing human-computer interaction scenarios, the model needs to treat different user groups fairly, but if the training data is incomplete or biased, it may lead to discriminatory output and loss of user trust.

In addressing these ethical issues, a multifaceted strategy can be adopted. First of all, in terms of data protection, encryption technology and data anonymization are adopted to ensure the security of user data during collection, transmission and storage. In addition, strict data use protocols are implemented to limit access to and use of sensitive data. Secondly, in order to reduce algorithm bias, the model should be trained using multi-subject data sets, and data enhancement techniques can also be used appropriately to improve the robustness of the model. Finally, ethical guidelines are proposed and followed to ensure ethical compliance in technology development and application. Through these measures, we can ensure the ethical and social responsibility of brain-inspired computing model while meeting the performance requirements of human-computer interaction.


To summarise, this section presents some current challenges and corresponding solutions for brain-inspired computational models applied to human-computer interaction. 
\section{FUTURE DIRECTIONS}

The previous chapter summarizes some challenges faced by HCI for brain-inspired computing models and gives some solutions. This chapter mainly summarizes possible future research directions. Inspired by the brain and the continuous progress and development of ML and DL models, the performance of HCI for brain-inspired computing tasks and the complexity of tasks that can be solved have been improved. In the future, more application scenarios can be broadened, BCI systems across language boundaries can be investigated, and paradigms for shifting neural network architectures can be transformed.
\label{section5}

\subsection{Landing Application of Brain-inspired Computing}
At present, brain-inspired computing models based on machine learning and deep learning have explored many tasks, such as binary classification tasks, information extraction tasks, and zero-sample emotion recognition tasks. One of the future research directions is to explore more application scenarios of brain-inspired computing. For example, in the fields of medicine and psychology, people with speech disorders can be helped to truly achieve silent electrical brain communication.

The research of human brain intelligence mechanism can also promote the innovation and expansion of artificial intelligence. Nowadays, neuromorphic chip, brain-computer interfaces, intelligent decision-making and other related research results are based on the extended development of brain science and computing science research, these studies still have a great space for development, and in the future may even subvert the current technical cognition.We should actively promote the product to create greater value for economic development and human social progress.

\subsection{BCIs across Language Boundaries}
BCI systems decode brain signals into text or speech to assist humans to interact with the outside world. Different languages have great differences in semantic understanding, semantic alignment and representation. Nowadays, the application of both open datasets and existing deep learning models in brain-inspired computing is mostly based on English. In fact, the application of BCI system should not be limited to English-speaking countries. For example, China is a large country with a population of 1.4 billion, and there is also a huge market demand for high-performance BCI systems. We hope that our suggestions can bring some inspiration to researchers to design a dedicated or even universal BCI system for humans of various languages in the world which will bring good news to more people suffering from diseases.

In addition, we can consider the brain signal as an intermediate bridge for language translation. For example, for the brain signal of the same thing, no matter what language the signal comes from, we can extract the semantic information carried by the brain signal itself, and then decode any required language. This research can enable a greater degree of cross-lingual BCI.

\subsection{Spiking Neural Network(SNN)}
In the past 10 years, artificial intelligence technology with deep neural networks as the main thrust has made great progress. Based on this, that is, the broad definition of brain-inspired computing, this paper summarizes the research progress of brain-inspired computing models based on deep learning and machine learning. In fact, as the most subtle intelligent organ in the human body, the human brain, only occupies 2\% to 3\% of the body weight, but almost controls all human behavior, so the performance of brain-inspired computing based on deep learning is far less than that of the human brain with about 100 billion neurons but very low consumption.

The human brain possesses the remarkable ability to self-organize and coordinate various cognitive functions, allowing for flexible adaptation to changing environments. A significant challenge in the realms of artificial intelligence and computational neuroscience is the integration of multi-scale biological principles to construct brain-inspired intelligent models\cite{zeng2023braincog}. SNN is a new generation of artificial neural network inspired by biology, which is oriented by brain science and develops along the direction of brain simulation. It expresses information flow with 0/1 pulse sequence, the encoding contains time information, and the internal neurons have dynamic characteristics, such as event-driven and sparse release. It converts input information into pulse sequence signals through pulse coding. And maintain the time relationship between pulses during the information transmission process, so neurons have microscopic memory properties\cite{nunes2022spiking,lagani2023spiking}. 

As the third generation of neural networks\cite{maass1997networks}, Spiking Neural Networks (SNNs) exhibit greater biological plausibility across multiple scales. This includes membrane potential, neuronal firing, synaptic transmission, synaptic plasticity, and the coordination of multiple brain areas. Crucially, SNNs offer enhanced biological interpret ability, increased energy efficiency, and a natural suitability for modeling diverse cognitive functions of the brain, making them well-suited for the creation of brain-inspired AI\cite{zeng2023braincog}.

In conclusion, the research on brain-inspired computing models applied to HCI is still in the preliminary stage, and there are many directions worth exploring in the future. Cross-lingual Bcis are more practical, while SNNs that simulate more biometric features are also a future trend.

\section{CONCLUSION}
\label{section6}
With the development of computer science, neuroscience, brain cognitive science and psychology, the acquisition and analysis of brain signals under various cognitive tasks has become an important intersection. Focusing on the human-computer interaction application scenarios of brain signal decoding text and speech, this study deeply discusses the brain-inspired computing models based on deep learning, summarizes the research status, progress and hotspots of brain-inspired computing for human-computer interaction, puts forward some existing limitations and gives some solutions. At present, this field is still in the initial stage of development. With the continuous evolution of artificial intelligence technology such as deep learning and the in-depth intersection of multi-disciplinary fields such as brain science, we believe that new breakthroughs can be continuously made. Therefore, we advocate continuing to pay attention to the co-evolution between multi-disciplines, and sincerely hope that this review can bring some inspiration to researchers in the field of human-computer interaction of brain-inspired computing and promote future innovation and progress.


\bibliographystyle{model1-num-names}

\bibliography{cas-refs}

\end{document}